\documentclass[11pt]{article}
\usepackage[preprint]{acl}
\usepackage{times}
\usepackage{latexsym}
\usepackage[T1]{fontenc}
\usepackage[utf8]{inputenc}
\usepackage{microtype}
\usepackage{inconsolata}
\usepackage{graphicx}
\usepackage{booktabs}
\usepackage{tabularx}
\usepackage{multirow}
\usepackage{amssymb}
\usepackage{fvextra}
\DefineVerbatimEnvironment{JSONExample}{Verbatim}{%
  fontsize=\footnotesize, frame=single, framesep=3pt,
  breaklines=true, breakanywhere=true, breakautoindent=true,
  breakindent=1em, breaksymbolleft={}}
\graphicspath{{figures/}}

\title{Right Values, Wrong Places: Diagnosing and Fixing Record Misplacement in LLM-Generated JSON}
\author{\textbf{Yiwei Zhang\textsuperscript{1,$\dagger$} \quad Chengke Wu\textsuperscript{2,$\dagger$} \quad Jiahao Li\textsuperscript{1}} \\
\textbf{Li Wang\textsuperscript{1} \quad Yao Yao\textsuperscript{3} \quad Jianqiang Li\textsuperscript{1,*}} \\
\textsuperscript{1}Shenzhen University \\
\textsuperscript{2}Shenzhen Institutes of Advanced Technology, Chinese Academy of Sciences \\
\textsuperscript{3}Chongqing University \\
\small $\dagger$ Equal contribution. \quad * Corresponding author.}

\begin{document}
\maketitle
\begin{abstract}
Large language models can generate valid JSON that preserves field values yet binds them to the wrong records or hierarchy levels. We introduce JSONPlacement, a controlled task supplying all values and assignment rules for transforming flat records into hierarchical JSON. Across six open-weight models, we identify two error types: record-assignment errors and hierarchy-realization errors. Whole-record moves account for 98.6\% of strictly traceable synthetic misplacements. We predict that aligning input with target traversal should reduce assignment errors, while replacing nested output with record-level paths should reduce hierarchy errors. With predictions fixed before generating new instances, we test both interventions separately and together. On synthetic instances, each nearly eliminates its targeted error while leaving the other. Combined, they achieve exact matches for three 8--12B models on all new synthetic instances and a real-data validation set. These findings motivate evaluating record placement explicitly and jointly designing input organization and output representation to preserve value--path bindings.

\end{abstract}

\section{Introduction}
\label{sec:introduction}

\begin{figure*}[t]
    \centering
    \includegraphics[width=\textwidth]{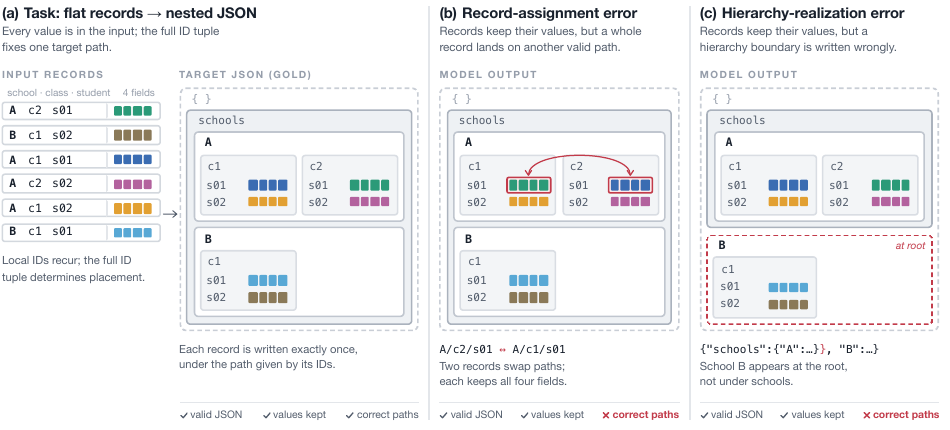}
    \caption{JSONPlacement and two error types. (a) Correct transformation. (b) Record-assignment error. (c) Hierarchy-realization error. Each four-cell color group represents one record. All three JSON structures are valid and preserve every value, but only (a) places all values correctly.}
    \label{fig:overview}
\end{figure*}

Large language models are used to write information into structured outputs, such as tables and hierarchical JSON, for downstream programs \citep{wu2022text,tang2024struc,agarwal2026paper}. A correct output must preserve field values. It must also place each record at the right position in the hierarchy. For example, a student in Class 1 may be written under Class 2. The name and grade are unchanged, but a downstream program now counts the student in the wrong class. Such an output can be valid JSON and keep every value, yet still express the wrong assignment.

Existing methods address different aspects of correctness. Constrained decoding restricts generation to a given grammar or schema \citep{geng2023grammar,chavan2026constrained}. A schema can rule out some hierarchy errors, but it can still permit a record under the wrong class. Fine-grained evaluation scores structure and content separately \citep{yang2026cspo}. It reveals failures that aggregate scores hide. Detecting an error, however, does not reveal where the error comes from. In end-to-end extraction, a model reads text and writes structured output in a single task \citep{lu2022uie,josifoski2022genie}. A misplaced value may then come from misreading the text or from misorganizing correct information. We therefore ask whether models place records correctly when all values and assignment rules are given. If they do not, we ask whether their errors follow reproducible patterns.

To answer these questions, we introduce JSONPlacement, a controlled task\footnote{Code is available at \url{https://anonymous.4open.science/r/artifact-75cf8a21/}; frozen tasks and raw model outputs are provided as supplementary data.}. The input is a list of flat records. Grouping fields determine where each record belongs. Content fields hold the values to preserve. The model must reorganize the records into JSON under a given hierarchy. For example, school, class, and student IDs determine where a student record belongs, and the name and grade are its content (Figure~\ref{fig:overview}a). A simple program can produce the correct output, so every instance has one unambiguous answer. This makes the task a controlled test of a step that LLM extraction pipelines perform implicitly: writing known records into a hierarchy. Because all values and assignments are given, a missing or misplaced record must come from organizing the output, not from reading or inference. We call the sequence of keys from the root to a record its record path. We trace where each input value appears in the model output.

Across models and structures, we find two main error types. A record-assignment error writes a record under another valid record path, such as a Class 1 student under Class 2 (Figure~\ref{fig:overview}b). The container hierarchy itself remains correct. A hierarchy-realization error changes the parent--child relations between containers. For example, a school is written at the root rather than inside the schools container (Figure~\ref{fig:overview}c). Both error types usually move whole records. Among strictly traceable misplaced values in synthetic tasks, 98.6\% move together with all other fields of their record.

The two error types suggest two testable interventions. When input records are shuffled, the model must find and gather records of the same group while generating. If this regrouping causes assignment errors, ordering the input to follow the target output should reduce them. This ordering does not change the target output. Hierarchy errors may arise because the model must keep nested containers open across many records. If so, writing each record as a separate entry keyed by its full path should reduce them. We call this a record-level path representation. The model must still derive each path from the grouping fields; we do not supply the answer paths. The representation changes the output form but keeps the same value--path bindings.

We fix our predictions before generating new instances. We then test each intervention alone and both together. On new synthetic instances, each intervention nearly eliminates its targeted error but leaves the other. Combined, they make three 8--12B models correct on every new synthetic instance across two target structures. With the same three models, the combination also makes every instance correct in a real-data validation. These results show that diagnosing placement errors can guide the joint design of input order and output representation.

Our contributions are threefold:
\begin{itemize}
    \item \textbf{Isolating placement failures.} We introduce a controlled task in which all information is given and every assignment is unambiguous. By tracing values and paths, we separate values that are kept but misplaced from values that are missing.
    \item \textbf{Characterizing error patterns.} We distinguish record-assignment errors from hierarchy-realization errors. We also measure how often strictly traceable synthetic misplacements move whole records.
    \item \textbf{Validating diagnosis-guided interventions.} We fix predictions before generating new instances. Synthetic validation shows that input ordering and record-level paths have complementary effects. Further tests cover larger models and real data.
\end{itemize}

\section{Related Work}
\label{sec:related-work}

\paragraph{Structured generation and organization.}
Generative extraction produces entities and relations \citep{lu2022uie,josifoski2022genie}, events \citep{lu2021text2event}, and code-style records \citep{li2023codeie}. G\&O separates answer generation from organization \citep{li2024simple}, while SLOT trains a JSON formatter \citep{wang2025slot}. Grammar constraints \citep{geng2023grammar,tuccio2025grammar}, schema and format training \citep{lu2025learning,wang2025verifiable}, and separating task solving from formatting \citep{deng2026decoupling} address output requirements. We instead control source records and assignments to diagnose placement.

\paragraph{Evaluation and error analysis.}
Constraints can improve logical parsing \citep{raspanti2025grammar} but hurt other tasks \citep{tam2024speak,schall2025hidden}; BoostCD combines constrained and unconstrained predictions \citep{sakota2025combining}. Structural compliance need not ensure semantic correctness \citep{chavan2026constrained}. CSPO evaluates structure and content \citep{yang2026cspo}; causal analysis assesses format effects \citep{yuan2026quantifying}. SLOT compares value similarity at matching nested keys, including ancestors \citep{wang2025slot}; AESOP optimally matches entity sets \citep{wu2024learning}; StructHallu-Drift studies schema-evolution errors \citep{hasan2026structhallu}. Our contribution is exact tracing of misplaced values, whole-record error analysis, and predictive intervention tests.

\paragraph{Order, position, and entity binding.}
Benchmarks also test how models interpret structured inputs and tool responses \citep{gu2025structext,kate2026good}. Output order can bias extraction \citep{li2023set}; input position affects retrieval and aggregation \citep{liu2024lost,hsieh2024ruler}, and key wording affects constrained generation \citep{le2026schema}. Record assignment relates to entity tracking and binding \citep{kim2023entity,feng2024binding,gurarieh2026mixing}. We connect output binding errors to testable changes in input order and output representation.

\section{JSONPlacement: Task Construction and Evaluation}
\label{sec:task}

\subsection{Task Definition}
\label{sec:task-definition}

JSONPlacement is a controlled task in which models transform flat records into hierarchical JSON. Each record contains grouping fields, which jointly determine the record's target path, and content fields, which supply the values to be written. Content-field names further determine the target leaf path for each value. Grouping identifiers may recur under different parents, so a local identifier alone is insufficient to determine assignment. All values and grouping rules are provided in the input, and a program deterministically generates the reference output, requiring neither external knowledge nor ambiguity resolution.

The prompt specifies target container names, hierarchical relationships, and a structural example. Each record must be written exactly once at its designated location, preserving the names, values, and types of its content fields, without omitting or adding fields, branches, or records. The target structure contains only objects and leaf values; array alignment is not involved. Appendix~\ref{app:examples} gives full input and output examples.

\subsection{Data Construction and Traceability}
\label{sec:data}

\paragraph{Synthetic tasks.}
Each instance contains 48 records with four content fields per record. We construct 24 instances for each of three structures: Original, Shallow--wide, and Deep--narrow; Table~\ref{tab:structures} specifies their grouping levels and branching factors. School-domain variants preserve record contents and tree shape to assess the effects of semantic-label changes.

\begin{table}[t]
    \centering
    \small
    \setlength{\tabcolsep}{3pt}
    \begin{tabular}{@{}lccc@{}}
        \toprule
        Structure & Levels & Branching factors & Records \\
        \midrule
        Original & 5 & $2\times3\times2\times2\times2$ & 48 \\
        Shallow--wide & 2 & $6\times8$ & 48 \\
        Deep--narrow & 5 & $2\times2\times2\times2\times3$ & 48 \\
        \bottomrule
    \end{tabular}
    \caption{Synthetic target structures. Levels count grouping fields rather than full JSON path depth.}
    \label{tab:structures}
\end{table}

\paragraph{Real records.}
We construct supplementary tasks from MBTA transit data and PDB protein-structure data, organizing preprocessed records into JSON according to given hierarchies. Models only reorganize the input, without retrieval, cross-table joins, or text extraction. Construction details and task sizes are given in Appendix~\ref{app:interventions}.

\paragraph{Traceability.}
The generator retains the correspondence among source fields, records, and target paths; synthetic content values are unique within each instance. A misplaced value is strictly traceable when its source is unique, it occurs once in the output, and it keeps its field name; a record moves whole when all of its content fields are traced to the same incorrect path prefix. Repeated values and output copies still count in scoring but are not used to identify moves. We report tracing coverage and leave untraced values unclassified.

\subsection{Evaluation Rules}
\label{sec:evaluation}

We compare the reference and model output by full paths, node types, and leaf values, ignoring object-key order and applying no output repair. Values are matched exactly by type and content after JSON decoding. Value preservation (VP) measures the fraction of target values retained regardless of path; value--path binding accuracy (VPA) additionally requires the correct path. Output precision measures the proportion of output leaf entries that are correctly bound, while whole-instance accuracy requires entirely correct structure and content.

VP, VPA, and output precision are macro-averaged over scorable instances, with the scorable count reported; truncated or invalid outputs count as whole-instance failures. Full metric definitions and scoring rules are provided in Appendix~\ref{app:scoring}.

\subsection{Experimental Protocol}
\label{sec:protocol}

We organize the evidence in three layers. Baseline evaluation and error analysis characterize the phenomenon and motivate explanations; validation on new instances, with predictions fixed in advance, tests the two main interventions; and real-record tasks examine the scope of the conclusions. Unique identifiers, local ordering, and constrained decoding serve only as diagnostics to test specific explanations.

The baseline uses shuffled inputs, identifiers that recur under different parents, and nested outputs. The prompt specifies only the target traversal order for the output and does not reorder input records. The baseline covers six models: Qwen3.5 at 4B, 9B, and 27B \citep{qwen2026qwen35}; Gemma 4 at 12B and 31B \citep{gemmateam2026gemma4}; and Ministral 3 8B \citep{liu2026ministral}. The core validation on new instances uses Qwen 9B, Gemma 12B, and Ministral 8B on 24 newly seeded instances each for Original and Deep--narrow. On the same instances, it compares the $2\times2$ conditions formed by shuffled versus aligned inputs and nested versus record-level path outputs. Predictions are fixed before generating new instances; all instances and outputs are scored, without selection, retries, or output repair.

Original has 24 base instances; Shallow--wide and Deep--narrow share a separate set of 24 base contents. Conditions sharing a base instance are analyzed as paired observations; neither models nor semantic variants increase the number of independent samples. Each batch retains tasks, prompts, reference outputs, raw model outputs, and generation parameters; cross-model differences are not attributed to parameter count alone.

\section{Right Values, Wrong Places}
\label{sec:errors}

\subsection{Values Survive, Placement Fails}
\label{sec:baseline}

\begin{table*}[t]
  \centering
  \small
  \setlength{\tabcolsep}{3pt}
  \renewcommand{\arraystretch}{1.12}
  \begin{tabular*}{\textwidth}{@{\extracolsep{\fill}}l rrr rrr rrr@{}}
    \toprule
    & \multicolumn{3}{c}{\textbf{Original}}
    & \multicolumn{3}{c}{\textbf{Shallow--wide}}
    & \multicolumn{3}{c}{\textbf{Deep--narrow}} \\
    \cmidrule(lr){2-4}\cmidrule(lr){5-7}\cmidrule(l){8-10}
    Model & VP (\%) & VPA (\%) & EM
          & VP (\%) & VPA (\%) & EM
          & VP (\%) & VPA (\%) & EM \\
    \midrule
    Qwen3.5-4B & 85.96 & 53.08 & 0/24 & 97.83 & 97.14 & 9/24 & 76.06 & 37.04 & 0/24 \\
    Qwen3.5-9B & 92.62 & 83.51 & 0/24 & 99.48 & 99.39 & 18/24 & 78.99 & 62.50 & 0/24 \\
    Qwen3.5-27B & 99.74 & 99.57 & 20/24 & 100.00 & 100.00 & 24/24 & 98.26 & 98.00 & 12/24 \\
    \addlinespace[5pt]
    Gemma 4 12B IT & 97.22 & 53.39 & 1/24 & 100.00 & 99.74 & 22/24 & 87.50 & 70.20 & 0/24 \\
    Gemma 4 31B IT & 99.91 & 99.91 & 23/24 & 100.00 & 100.00 & 24/24 & 98.35 & 98.00 & 12/24 \\
    \addlinespace[5pt]
    Ministral 3 8B Instruct & 95.83 & 66.49 & 0/24 & 99.13 & 98.96 & 15/24 & 86.81 & 44.88 & 0/24 \\
    \bottomrule
  \end{tabular*}
  \par\vspace{4pt}
  \begin{minipage}{\textwidth}
    \footnotesize
    VP: value preservation; VPA: value--path binding accuracy; EM: exact-match count out of 24.
    VP/VPA are macro-averaged over scorable outputs; truncated or invalid outputs count as EM failures.
  \end{minipage}
  \caption{Baseline performance across six models and three target structures.}
  \label{tab:baseline-main}
\end{table*}

\begin{figure}[t]
    \centering
    \includegraphics[width=\columnwidth]{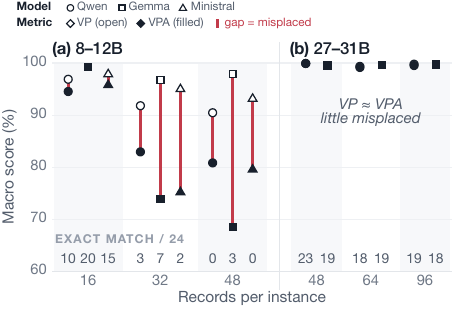}
    \caption{Baseline VP and VPA on Original as records per instance increase. Red bars mark values that are preserved but misplaced; numbers above the axis give exact matches out of 24. Each record count uses separately generated instances.}
    \label{fig:gradient}
\end{figure}

The central observation is that models often keep the values but put them in the wrong places. On Original, Gemma 12B preserves 97.2\% of target values yet binds only 53.4\% to the correct path, and no 8--12B model solves more than one of 24 instances (Table~\ref{tab:baseline-main}). Most failures thus come from misplaced values rather than missing ones; only on Deep--narrow do smaller models also lose substantial content (VP 76--79\% for Qwen 4B and 9B).

The gap emerges as instances grow. On Original with 16, 32, and 48 records per instance (Figure~\ref{fig:gradient}a), VP and VPA nearly coincide at 16 records, where 10--20 of 24 instances are solved. At 32 records, VP stays above 91\% while VPA falls to 74--83\%, and no model solves more than 7 instances. Beyond this point the change is not monotone; Ministral's VPA partly recovers at 48 records.

Larger models narrow the gap but do not close it. On Original, Qwen 27B and Gemma 31B keep VP and VPA nearly equal up to 96 records (Figure~\ref{fig:gradient}b); their remaining failures stem from a handful of omitted or misplaced records. Both also nearly saturate Shallow--wide, yet each solves only 12 of 24 Deep--narrow instances (Table~\ref{tab:baseline-main}). As the structures differ in both depth and branching, and the larger models ran in a separate batch, we attribute these differences neither to depth nor to scale alone. Overall, value preservation and correct placement come apart; \S\ref{sec:record-assignment}--\ref{sec:hierarchy-realization} decompose this gap into two error types.

\subsection{Record-Assignment Errors}
\label{sec:record-assignment}

Traceable synthetic misplacements predominantly involve whole records. When a record is assigned to a wrong path, its content fields often move together to another valid path, typically one that differs by a single grouping identifier (Figure~\ref{fig:overview}b). In four earlier synthetic conditions, 80.8\% of misplaced leaves can be traced to a unique destination, and 98.6\% of these belong to whole-record moves. The remaining leaves are not classified; even if all of them were isolated errors, whole-record moves would still account for 79.7\% of all misplaced leaves (Appendix~\ref{app:record-assignment}). Were fields to err independently, only about 0.06 records would be expected to move whole; we observe 6,293.

The pattern persists across the tested record sizes and checking instructions. In a field-count control with 2, 4, or 8 fields per record, 653 of 654 erroneous records still move as a unit. Asking models to verify each record's full path before answering leaves the pattern intact. The path check does not significantly improve VPA over the general check for either model (Appendix~\ref{app:record-assignment}).

When a record is absent from the entire output, its target location is often occupied by another whole record (91\% for Qwen 9B, 49\% for Gemma 12B). Occupants typically differ from the intended record by one grouping identifier (86\% and 91\%, versus about 12\% at random). Whether the two records are adjacent in the input has little effect.

Records are also omitted or duplicated whole, and omissions concentrate early in the input. In the two checks above, the first 20\% of input records disappear 4.3 (Qwen) and 2.9 (Gemma) times as often as the rest. Since input order is randomized, this effect reflects position rather than content.

To test whether path similarity contributes to these confusions, we make every grouping identifier unique within an instance, so that distinct paths no longer share identifiers (setup and full results in Appendix~\ref{app:record-assignment}). On new instances, records written to other valid paths drop from 111 to 8 (Qwen) and from 91 to 36 (Ministral) on Original, and to zero for all three models on Deep--narrow. Whole-instance accuracy, however, barely improves, because whole-record omissions and out-of-tree records remain.

These analyses identify whole-record movement between similar paths as a dominant pattern of record-assignment errors. Because removing path similarity changes the task itself, Section~\ref{sec:interventions} instead tests an intervention that leaves the task intact, namely ordering the input by target traversal.

\begin{table*}[t]
  \centering
  \small
  \setlength{\tabcolsep}{2.5pt}
  \renewcommand{\arraystretch}{1.12}
  \begin{tabular*}{\textwidth}{@{\extracolsep{\fill}}l cc rr@{\hspace{6pt}}rrr rr@{\hspace{6pt}}rrr@{}}
    \toprule
    & & & \multicolumn{5}{c}{\textbf{Original}} & \multicolumn{5}{c}{\textbf{Deep--narrow}} \\
    \cmidrule(lr){4-8}\cmidrule(l){9-13}
    & & & & & \multicolumn{3}{c}{Record errors} & & & \multicolumn{3}{c}{Record errors} \\
    \cmidrule(lr){6-8}\cmidrule(l){11-13}
    Model & Order & Path & EM & VPA (\%) & Other & Outside & Omitted & EM & VPA (\%) & Other & Outside & Omitted \\
    \midrule
    \multirow{4}{*}{Qwen3.5-9B}
      & -- & -- & 0/24 & 80.82 & 111 & 0 & 110 & 0/24 & 60.33 & 121 & 3 & 286 \\
      & \checkmark & -- & 24/24 & 100.00 & 0 & 0 & 0 & 23/24 & 99.48 & 6 & 0 & 0 \\
      & -- & \checkmark & 0/24 & 80.42 & 84 & 0 & 104 & 0/24 & 72.14 & 107 & 0 & 103 \\
      & \checkmark & \checkmark & 24/24 & 100.00 & 0 & 0 & 0 & 24/24 & 100.00 & 0 & 0 & 0 \\
    \addlinespace[5pt]
    \multirow{4}{*}{Gemma 4 12B IT}
      & -- & -- & 3/24 & 68.49 & 5 & 333 & 25 & 0/24 & 62.95 & 86 & 147 & 175 \\
      & \checkmark & -- & 8/24 & 66.32 & 0 & 384 & 4 & 2/24 & 56.25 & 0 & 504 & 0 \\
      & -- & \checkmark & 12/24 & 97.74 & 10 & 0 & 14 & 1/24 & 76.43 & 80 & 23 & 98 \\
      & \checkmark & \checkmark & 24/24 & 100.00 & 0 & 0 & 0 & 24/24 & 100.00 & 0 & 0 & 0 \\
    \addlinespace[5pt]
    \multirow{4}{*}{Ministral 3 8B Instruct}
      & -- & -- & 0/24 & 79.62 & 91 & 59 & 75 & 0/24 & 54.08 & 197 & 214 & 118 \\
      & \checkmark & -- & 17/24 & 78.12 & 0 & 252 & 0 & 2/24 & 51.04 & 0 & 564 & 0 \\
      & -- & \checkmark & 0/24 & 84.31 & 100 & 1 & 64 & 0/24 & 79.96 & 149 & 1 & 68 \\
      & \checkmark & \checkmark & 24/24 & 100.00 & 0 & 0 & 0 & 24/24 & 100.00 & 0 & 0 & 0 \\
    \bottomrule
  \end{tabular*}
  \par\vspace{4pt}
  \begin{minipage}{\textwidth}
    \footnotesize
    Order: inputs ordered by target traversal; Path: record-level paths. EM: exact-match count out of 24.
    Record errors: whole records written to another valid path (Other), outside the target tree (Outside), or omitted (Omitted),
    summed over scorable outputs. VPA is macro-averaged over scorable outputs; truncated or invalid outputs count as EM failures.
  \end{minipage}
  \caption{Validation on new instances for the two interventions and their combination.}
  \label{tab:interventions}
\end{table*}

\subsection{Hierarchy-Realization Errors}
\label{sec:hierarchy-realization}

The second error form preserves field values but breaks hierarchy boundaries, so whole records fall outside the target tree (Figure~\ref{fig:overview}c). On Original under the baseline, Gemma writes 497 records outside the target tree and only 8 to other valid paths; Qwen writes none outside the tree. Gemma thus keeps nearly all values yet places only about half correctly (\S\ref{sec:baseline}).

These records arise from two closure errors. In premature closure, a container ends too early and later groups are written at a shallower level. In under-closure, a container is left open and the next sibling is nested inside it. Among these synthetic closure cases, premature closure occurs only in Gemma, whereas Ministral exhibits under-closure.

To test the explanation based on container closure, we constrain decoding with a grammar of the target tree shape. It fixes the container nesting, key names, and identifier formats stated in the prompt, but no record bindings or values (setup and full results in Appendix~\ref{app:hierarchy}). Inputs follow target order to reduce assignment errors while testing closure. Requiring only valid JSON leaves both closure errors nearly unchanged. With the tree-shape grammar, Ministral is correct on every instance. Gemma no longer closes containers early, but most such instances now end with an entire top-level group missing; additionally fixing the number of top-level groups makes it correct on every instance as well. No constrained output places a record on another valid path. Gemma thus appears to treat its output as complete where it would otherwise close early.

These diagnostics connect the observed synthetic hierarchy failures to container closure. Out-of-tree placement alone does not identify a closure error, since incorrect keys can produce the same destination. Section~\ref{sec:interventions} tests interventions for record assignment and container closure separately and together.

\section{Targeted Interventions}
\label{sec:interventions}

Section~\ref{sec:errors} suggests different sources for the two error types, and we test one intervention for each. Ordering input records by target traversal targets record assignment. If assignment errors arise when the model must regroup records by path because the input does not follow the target order, ordering should reduce them. Record-level paths target hierarchy realization. If hierarchy errors arise from keeping nested containers open across records, writing each record under its full path prefix should reduce them. Both interventions keep the records and their target positions fixed. Before generating new instances, we predicted that each intervention would reduce its targeted error type but not the other, and that their combination would perform best.

\subsection{Complementary Effects on New Synthetic Instances}
\label{sec:complementary}

On new synthetic instances, each intervention nearly eliminates its targeted error type while leaving substantial errors of the other type, consistent with the predictions (Table~\ref{tab:interventions}). Ordering nearly eliminates records written to other valid paths and whole-record omissions for all three models, and Qwen solves 47 of 48 instances. Gemma and Ministral, however, still fail on most instances, as their out-of-tree records persist and even increase. Ordering thus largely resolves assignment errors in this synthetic validation, but not closure errors.

Record-level paths show the opposite pattern. Out-of-tree records nearly disappear, yet many records are still written to other valid paths for all three models, and exact matches rise substantially only for Gemma on Original. They also lengthen outputs, and more outputs reach the token limit (setup and full results in Appendix~\ref{app:interventions}). Unlike the tree-shape grammar in \S\ref{sec:hierarchy-realization}, they need neither a decoding constraint nor the number of top-level groups in advance.

Combining the two, all three models solve every instance of both structures, with no misplaced or omitted records.

To test whether ordering helps by removing the need to regroup, we reuse the earlier Original instances and shuffle target-ordered records only within blocks of eight, so that the input departs from target order only locally and regrouping spans fewer records (setup and full results in Appendix~\ref{app:interventions}). For Qwen, exact matches rise from 0 under full shuffling to 10 under local shuffling and 24 under full ordering; Gemma moves only from 2 to 3 to 7, consistent with its errors being mainly hierarchical. For the model dominated by assignment errors, the gain thus grows as less regrouping is required.

Overall, the interventions have complementary effects on the two error types in this synthetic validation, and their combination yields entirely correct outputs.

\subsection{Validation Across Larger Models and Real Data}
\label{sec:dominant-error}

\begin{table}[t]
  \centering
  \small
  \setlength{\tabcolsep}{2.5pt}
  \renewcommand{\arraystretch}{1.12}
  \begin{tabular*}{\columnwidth}{@{\extracolsep{\fill}}l rrr rrrr@{}}
    \toprule
    & \multicolumn{3}{c}{\textbf{Baseline errors}} & \multicolumn{4}{c}{\textbf{EM}} \\
    \cmidrule(lr){2-4}\cmidrule(l){5-8}
    Model & Oth. & Out. & Omit. & Base & Order & Path & Both \\
    \midrule
    \multicolumn{8}{@{}l}{\textit{Synthetic, 27--31B (48/64/96 records, 24 each)}} \\
    Qwen3.5-27B    & 9  & 0     & 16 & 60 & 72 & 53 & 72 \\
    Gemma 4 31B & 1  & 0     & 17 & 56 & 72 & 62 & 72 \\
    \addlinespace[5pt]
    \multicolumn{8}{@{}l}{\textit{PDB, 8--12B (48 instances)}} \\
    Qwen3.5-9B     & 8  & 14    & 22 & 18 & 48 & 11 & 48 \\
    Gemma 4 12B & 0  & 29    & 26 & 22 & 48 & 17 & 48 \\
    Ministral 3 8B & 31 & 1,992 & 8  & 7  & 45 & 33 & 48 \\
    \bottomrule
  \end{tabular*}
  \par\vspace{4pt}
  \begin{minipage}{\columnwidth}
    \footnotesize
    Columns and conditions as in Table~\ref{tab:interventions}; baseline errors are record counts summed over scorable baseline outputs.
    Truncated or invalid outputs count as EM failures.
  \end{minipage}
  \caption{Gains of each intervention for larger models and real records.}
  \label{tab:dominant-error}
\end{table}

We next test whether the intervention benefits extend to larger models and real records. We extend Qwen 27B and Gemma 31B to 64 and 96 records under the same four conditions (Table~\ref{tab:dominant-error}; setup and per-level results in Appendix~\ref{app:interventions}). At every level, their failures come from a few records omitted or misplaced as a whole, and no record is written outside the target tree. Ordering removes these errors and, like the combination, solves every instance. Record-level paths alone raise Gemma 31B's total exact matches from 56 to 62 out of 72, but lower Qwen 27B's from 60 to 53. Here, input ordering suffices to resolve the remaining record-level failures while preserving nested output.

Real records further test whether the synthetic error patterns and intervention effects carry over. We build 48 tasks from PDB protein atom records, grouped by model, chain, residue, and atom, with 48--66 records per task (construction and sample in Appendix~\ref{app:interventions}). For Qwen and Gemma, baseline failures involve few records, scattered across omissions and misplacements. Record-level paths alone lower their exact matches, whereas ordering solves every instance. Ministral preserves nearly all values, yet most of its records fall outside the target tree because atoms are keyed by serial numbers rather than by atom names. This is a key-selection error, not a container-closure error, although both can produce out-of-tree records. Record-level paths alone remove most such records and raise exact matches substantially. Unlike in the synthetic closure cases, ordering also removes most of these errors. Ordering's effect on key selection extends its benefit beyond the synthetic assignment cases. With both interventions, all three models solve every instance.

MBTA transit tasks test ordering only, asking whether it still resolves assignment when paths become harder to distinguish. Records are grouped by route, direction, trip, and stop, with 48 records per task (construction and full results in Appendix~\ref{app:interventions}). We replace trip IDs within each route and direction with local labels such as \texttt{t1} and \texttt{t2}, so that the same label recurs under different routes and directions. Baseline exact matches drop for both Qwen and Gemma, and a substantial share of the added errors, especially for Gemma, are records written or copied under a same-named trip of another route or direction. Ordering solves every instance for both models with both original and local IDs. Consistent with \S\ref{sec:record-assignment}, less distinguishable paths increase assignment errors, and ordering removes them.

Across these extensions, ordering improves record placement while preserving nested output. Record-level paths reduce synthetic closure errors and real-data key-selection errors. Their combination solves every instance in the core synthetic validation, the larger-model extension, and PDB. The MBTA ordering tests further support organizing inputs by target traversal when local identifiers repeat.

\section{Discussion}
\label{sec:discussion}

\paragraph{Evaluate placement, not only values.}
Both failure modes keep every value and pass syntax checks. Evaluations of structured generation should therefore trace value--path bindings. Telling the two modes apart also matters, because they call for different fixes.

\paragraph{Diagnose before intervening.}
Our interventions were derived from error patterns, not found by trial and error. The same workflow applies to other structured outputs. Trace where errors land, hypothesize the generation demand they imply, and test a change that removes only that demand.

\paragraph{Design inputs and outputs around the diagnosed demand.}
Assignment errors call for changes to input organization. Records of the same group should appear together. In a two-stage pipeline, we suggest sorting generated records by their grouping fields before organization. Hierarchy errors call for changes to output representation. Stating each record's full path avoids long-lived nesting, and a program can assemble the hierarchy. When both modes occur, both sides need to change.

\section{Conclusion}
\label{sec:conclusion}

We introduced JSONPlacement, a controlled task in which all values and assignment rules are given. It reveals two failure modes in LLM-generated JSON: record-assignment errors and hierarchy-realization errors. Both keep values and usually move whole records. We linked each mode to a hypothesized demand of generation and fixed predictions before testing. Input ordering sharply reduced assignment errors, while record-level paths sharply reduced hierarchy errors. Combined, they made every instance correct in the core synthetic validation, the larger-model extension, and the real-data validation of both interventions. These findings suggest a workflow for structured generation. Evaluate value--path bindings, diagnose the dominant failure mode, and change the input or output to remove its demand.

\section*{Limitations}

\paragraph{Task scope and transfer.}
The task supplies complete records and explicit grouping rules, so it does not measure the contribution of placement failures to end-to-end extraction errors. Exact tracing requires known target bindings and identifiable source records; applying it to extraction would require additional annotation or alignment. Input ordering applies only once grouping fields are explicit. Whether the intervention benefits transfer to tasks where assignments must first be inferred from text remains untested. Real-data validation covers only two sources, and MBTA tests ordering alone; the combined intervention has therefore not been validated across multiple real-data sources.

\paragraph{Scope of the explanations.}
Our evidence establishes behavioral patterns, not a unique internal mechanism. Record-level paths change output length, bracket structure, and grouping-key repetition together, so their effects cannot be attributed to any single factor. Distinguishing these explanations requires experiments that vary these factors independently.

\bibliography{references}

\appendix
\section{Scoring Details}
\label{app:scoring}

Table~\ref{tab:metrics} defines the evaluation metrics.

\begin{table}[htbp]
    \centering
    \small
    \setlength{\tabcolsep}{4pt}
    \renewcommand{\arraystretch}{1.12}
    \begin{tabularx}{\columnwidth}{@{}>{\raggedright\arraybackslash}p{0.35\columnwidth}>{\raggedright\arraybackslash}X@{}}
        \toprule
        Metric & Definition \\
        \midrule
        VP & Target leaves matched one-to-one to output values, ignoring paths / all target leaves \\
        \addlinespace[3pt]
        VPA & Target leaves with the correct value and type at the target path / all target leaves \\
        \addlinespace[3pt]
        Output precision & Correctly bound target leaves / all raw output leaf entries, including duplicates \\
        \addlinespace[3pt]
        Whole-instance accuracy & Entirely correct outputs / instances with model outputs; no missing or extra paths, duplicate declarations, or conflicts \\
        \bottomrule
    \end{tabularx}
    \caption{Evaluation metrics.}
    \label{tab:metrics}
\end{table}

The record-level path representation uses each full record path as a key and the record's field object as its value. The evaluator interprets the target tree according to the paths actually declared, without using the reference to correct addresses, complete records, or repair the hierarchy.

Values are compared exactly by type and content after JSON decoding, without semantic normalization; for example, the string \texttt{"123"} differs from the number \texttt{123}. Matching first preserves correct value--path bindings, then matches remaining equal-valued entries. No output value can satisfy multiple targets. High VP with low VPA thus indicates that most values remain but are not correctly bound. High VPA does not guarantee an entirely correct output either, because extra branches or duplicate declarations may remain.

The evaluator retains every declaration of duplicate keys rather than keeping only the first or last. Repeated declarations of the same correct value at one path contribute at most one correct binding, but count as redundancy and prevent whole-instance correctness. If a path declares both a correct and an incorrect value, it receives no correct-binding credit. Conflicts between values and containers likewise prevent the affected paths from being counted as correct.

VP, VPA, and output precision are macro-averaged over scorable instances, with the scorable count reported. Truncation, invalid JSON, and uninterpretable representations count as whole-instance failures but are excluded from complete-output metrics. API failures are reported separately and excluded from the whole-instance accuracy denominator, which covers instances with model outputs. When comparing conditions, whole-instance correct counts retain failed outputs, whereas VP, VPA, and related differences use only paired instances scorable under both conditions. Error diagnosis uses the same scoring and tracing results rather than a separate correctness measure. Labels such as omission, duplication, conflict, and hierarchy displacement can overlap and must not be summed as mutually exclusive categories.

\section{Task Examples}
\label{app:examples}

\begin{figure*}[p]
  \centering
  \begin{minipage}[t]{0.485\textwidth}
    \footnotesize
    \textbf{Source JSON (all four records)}
\begin{JSONExample}
{
 "address_records": [
  {
   "tenant_id": "t17",
   "order_id": "o30",
   "shipment_id": "s41",
   "package_id": "p22",
   "address_role": "sender",
   "city": "Spokane",
   "street": "813 Maple Street",
   "postal_code": "23605",
   "contact_name": "David Morgan"
  },
  {
   "tenant_id": "t17",
   "order_id": "o30",
   "shipment_id": "s41",
   "package_id": "p22",
   "address_role": "recipient",
   "city": "Pittsburgh",
   "street": "303 Maple Street",
   "postal_code": "26482",
   "contact_name": "Alice Ellis"
  },
  {
   "tenant_id": "t81",
   "order_id": "o30",
   "shipment_id": "s41",
   "package_id": "p22",
   "address_role": "sender",
   "city": "Detroit",
   "street": "1085 Birch Road",
   "postal_code": "26893",
   "contact_name": "Elena Dawson"
  },
  {
   "tenant_id": "t81",
   "order_id": "o30",
   "shipment_id": "s41",
   "package_id": "p22",
   "address_role": "recipient",
   "city": "Toledo",
   "street": "949 Cedar Lane",
   "postal_code": "24975",
   "contact_name": "Bruno Grant"
  }
 ]
}
\end{JSONExample}
  \end{minipage}\hfill
  \begin{minipage}[t]{0.485\textwidth}
    \footnotesize
    \textbf{Complete reference output}
\begin{JSONExample}
{
 "tenants": {
  "t17": {"orders": {
   "o30": {"shipments": {
    "s41": {"packages": {
     "p22": {"addresses": {
      "sender": {
       "city": "Spokane",
       "street": "813 Maple Street",
       "postal_code": "23605",
       "contact_name": "David Morgan"
      },
      "recipient": {
       "city": "Pittsburgh",
       "street": "303 Maple Street",
       "postal_code": "26482",
       "contact_name": "Alice Ellis"
      }
     }}
    }}
   }}
  }},
  "t81": {"orders": {
   "o30": {"shipments": {
    "s41": {"packages": {
     "p22": {"addresses": {
      "sender": {
       "city": "Detroit",
       "street": "1085 Birch Road",
       "postal_code": "26893",
       "contact_name": "Elena Dawson"
      },
      "recipient": {
       "city": "Toledo",
       "street": "949 Cedar Lane",
       "postal_code": "24975",
       "contact_name": "Bruno Grant"
      }
     }}
    }}
   }}
  }}
 }
}
\end{JSONExample}
  \end{minipage}
  \caption{An Original example: four source records (left) and their reference output (right).}
  \label{fig:example-original}
\end{figure*}

\begin{figure*}[p]
  \centering
  \begin{minipage}[t]{0.485\textwidth}
    \footnotesize
    \textbf{Record-level paths: four entries}
\begin{JSONExample}
{
 "/tenants/t17/orders/o30/shipments/s41/packages/p22/addresses/sender": {
  "city": "Spokane",
  "street": "813 Maple Street",
  "postal_code": "23605",
  "contact_name": "David Morgan"
 },
 "/tenants/t17/orders/o30/shipments/s41/packages/p22/addresses/recipient": {
  "city": "Pittsburgh",
  "street": "303 Maple Street",
  "postal_code": "26482",
  "contact_name": "Alice Ellis"
 },
 "/tenants/t81/orders/o30/shipments/s41/packages/p22/addresses/sender": {
  "city": "Detroit",
  "street": "1085 Birch Road",
  "postal_code": "26893",
  "contact_name": "Elena Dawson"
 },
 "/tenants/t81/orders/o30/shipments/s41/packages/p22/addresses/recipient": {
  "city": "Toledo",
  "street": "949 Cedar Lane",
  "postal_code": "24975",
  "contact_name": "Bruno Grant"
 }
}
\end{JSONExample}
  \end{minipage}\hfill
  \begin{minipage}[t]{0.485\textwidth}
    \footnotesize
    \textbf{Leaf-level paths: sixteen entries}
\begin{JSONExample}
{
 "/tenants/t17/orders/o30/shipments/s41/packages/p22/addresses/sender/city": "Spokane",
 "/tenants/t17/orders/o30/shipments/s41/packages/p22/addresses/sender/street": "813 Maple Street",
 "/tenants/t17/orders/o30/shipments/s41/packages/p22/addresses/sender/postal_code": "23605",
 "/tenants/t17/orders/o30/shipments/s41/packages/p22/addresses/sender/contact_name": "David Morgan",
 "/tenants/t17/orders/o30/shipments/s41/packages/p22/addresses/recipient/city": "Pittsburgh",
 "/tenants/t17/orders/o30/shipments/s41/packages/p22/addresses/recipient/street": "303 Maple Street",
 "/tenants/t17/orders/o30/shipments/s41/packages/p22/addresses/recipient/postal_code": "26482",
 "/tenants/t17/orders/o30/shipments/s41/packages/p22/addresses/recipient/contact_name": "Alice Ellis",
 "/tenants/t81/orders/o30/shipments/s41/packages/p22/addresses/sender/city": "Detroit",
 "/tenants/t81/orders/o30/shipments/s41/packages/p22/addresses/sender/street": "1085 Birch Road",
 "/tenants/t81/orders/o30/shipments/s41/packages/p22/addresses/sender/postal_code": "26893",
 "/tenants/t81/orders/o30/shipments/s41/packages/p22/addresses/sender/contact_name": "Elena Dawson",
 "/tenants/t81/orders/o30/shipments/s41/packages/p22/addresses/recipient/city": "Toledo",
 "/tenants/t81/orders/o30/shipments/s41/packages/p22/addresses/recipient/street": "949 Cedar Lane",
 "/tenants/t81/orders/o30/shipments/s41/packages/p22/addresses/recipient/postal_code": "24975",
 "/tenants/t81/orders/o30/shipments/s41/packages/p22/addresses/recipient/contact_name": "Bruno Grant"
}
\end{JSONExample}
  \end{minipage}
  \caption{The example of Figure~\ref{fig:example-original} in record-level paths (left) and leaf-level paths (right).}
  \label{fig:example-paths}
\end{figure*}

This appendix illustrates the task in \S\ref{sec:task-definition}. All displayed outputs are reference answers, not model outputs.

\subsection{Input and Output Format}
\label{app:example-contract}

Each source record has grouping fields and four content fields: \texttt{city}, \texttt{street}, \texttt{postal\_code}, and \texttt{contact\_name}. The full tuple of grouping fields determines one record path, and appending a field name determines one target leaf. All content values are strings and must be copied exactly. In the nested output, each container maps entity identifiers to objects that hold the next container, and grouping fields appear only as keys along the path. Every record must appear exactly once, with no extra branches or fields. Evaluation ignores key order.

Figure~\ref{fig:example-original} shows four records from an Original instance and their reference output. The identifiers \texttt{o30/s41/p22} recur under tenants \texttt{t17} and \texttt{t81}, so a record's place depends on all five grouping fields, not on the lower identifiers alone.

\subsection{Structures and Domains}
\label{app:example-schemas}

Table~\ref{tab:example-schemas} lists the grouping fields of each structure in both domains. Shallow--wide writes the content fields directly in each order; Deep--narrow replaces the sender and recipient roles of Original with address identifiers. The school version renames grouping fields, containers, and roles, but keeps identifier strings, content values, record order, and tree shape.

\begin{table}[t]
  \centering
  \small
  \setlength{\tabcolsep}{3pt}
  \renewcommand{\arraystretch}{1.12}
  \begin{tabularx}{\columnwidth}{@{}l >{\raggedright\arraybackslash}X >{\raggedright\arraybackslash}X@{}}
    \toprule
    Structure & Logistics & School \\
    \midrule
    Original & tenant, order, shipment, package, role (sender/recipient) & school, campus, class, student, role (student/guardian) \\
    \addlinespace[3pt]
    Shallow--wide & tenant, order & school, student \\
    \addlinespace[3pt]
    Deep--narrow & tenant, order, shipment, package, address & school, campus, grade, class, student \\
    \bottomrule
  \end{tabularx}
  \caption{Grouping fields by structure and domain, from the top level down. Each field has a corresponding plural container, such as \texttt{tenants}.}
  \label{tab:example-schemas}
\end{table}

\begin{table}[t]
  \centering
  \small
  \setlength{\tabcolsep}{2pt}
  \renewcommand{\arraystretch}{1.12}
  \begin{tabular*}{\columnwidth}{@{\extracolsep{\fill}}l l l l@{}}
    \toprule
    Condition & Input & IDs & Output \\
    \midrule
    Baseline           & Shuffled & Repeated & Nested \\
    Aligned            & Target   & Repeated & Nested \\
    Unique IDs         & Shuffled & Unique           & Nested \\
    Leaf-level paths & Shuffled & Repeated & Leaf paths \\
    Record-level paths & Shuffled & Repeated & Record paths \\
    \bottomrule
  \end{tabular*}
  \caption{What each condition changes. All conditions keep the records and tree shape; only unique identifiers change the target paths.}
  \label{tab:example-conditions}
\end{table}

\subsection{Conditions and Path Representations}
\label{app:example-conditions}

Table~\ref{tab:example-conditions} lists what each condition changes. Figure~\ref{fig:example-paths} writes the Original example in the two path representations. Record-level paths map each record's full path to its content object, giving four entries; leaf-level paths add the field name to the key, giving sixteen. Both encode the same sixteen value--path bindings as the nested reference. Keys follow the JSON Pointer syntax \citep{bryan2013jsonpointer}.

\section{Record-Assignment Analysis Details}
\label{app:record-assignment}

This appendix gives the scope, tracing rules, and full results behind \S\ref{sec:record-assignment}. The pooled audit, the field-count and checking controls, and the unique-identifier diagnostic use different outputs and are reported separately.

\subsection{Whole-Record Movement}
\label{app:record-pool}

The 98.6\% figure pools four conditions from the earlier synthetic runs: the baseline, aligned input, unique identifiers, and leaf-level path output. Table~\ref{tab:record-pool} lists the 1,728 outputs, of which 1,529 are scorable. Conditions, models, and domains reuse the same base contents, so these outputs are repeated observations rather than independent instances.

\begin{table}[t]
  \centering
  \small
  \setlength{\tabcolsep}{4pt}
  \renewcommand{\arraystretch}{1.12}
  \begin{tabular*}{\columnwidth}{@{\extracolsep{\fill}}l rr rr@{}}
    \toprule
    & \multicolumn{2}{c}{\textbf{Original}} & \multicolumn{2}{c}{\textbf{Shallow + Deep}} \\
    \cmidrule(lr){2-3}\cmidrule(l){4-5}
    Model & Logistics & School & Logistics & School \\
    \midrule
    Qwen 9B      & 88/96 & 84/96 & 176/192 & 180/192 \\
    Gemma 12B    & 94/96 & 91/96 & 181/192 & 184/192 \\
    Ministral 8B & 95/96 & 95/96 & 188/192 & -- \\
    OLMo 7B      & --    & --    & 73/192  & -- \\
    \midrule
    Total & \multicolumn{4}{r}{1,529/1,728} \\
    \bottomrule
  \end{tabular*}
  \caption{Scorable outputs in the pooled audit. Each cell covers four conditions with 24 instances per structure.}
  \label{tab:record-pool}
\end{table}

A misplaced value is strictly traceable when its source is unique, it occurs once in the output, and it keeps its field name. A record moves whole when all four of its fields appear under the same incorrect record prefix. Of the 31,577 misplaced leaves, 25,519 (80.8\%) are strictly traceable, and the 6,293 whole-record moves account for 25,172 of them (98.6\%). The 6,058 leaves outside the strict scope are not counted as isolated field errors. Even if all of them were, whole-record moves would still account for 25,172 of the 31,577 misplaced leaves (79.7\%). Of the 6,293 moves, 1,485 land on another valid record path and 4,808 fall outside the target tree, so the figure describes the granularity of both error forms.

The reference for independent fields permutes, within each instance, the destination prefixes of each field across source records, keeping each field's misplacement count and destination distribution. For $n$ records with $k$ fields, let $c_j(d)$ count field $j$ assigned to incorrect prefix $d$, and let $p_r$ be the target prefix of record $r$. The expected number of whole-record moves is
\[
  E[X]=\sum_{r=1}^{n}\sum_{d\ne p_r}\prod_{j=1}^{k}\frac{c_j(d)}{n},
\]
summed over instances. This gives 0.057 expected moves against 6,293 observed; by Markov's inequality, $P(X\ge6{,}293)\le9.0\times10^{-6}$.

\subsection{Field-Count and Checking Controls}
\label{app:record-controls}

The field-count control fixes 192 target fields per instance and varies their split: 96 records with two fields, 48 with four, or 24 with eight. Each configuration uses 24 new instances of the five-level logistics hierarchy, with shuffled inputs and nested outputs, for Qwen 9B and Gemma 12B. The checking control reuses these 72 instances and adds one instruction to the prompt. The general check asks the model to confirm that every record and field is present and copied exactly; the path check asks it to read each record's full identifier tuple and place all its fields under the matching path. Neither asks the model to print the check.

A record enters the analysis when each of its fields is either correctly bound or strictly traceable to an incorrect prefix. Among such records with at least one misplaced field, nearly all move whole (Table~\ref{tab:record-controls}): 653 of 654 under the field-count control and 785 of 786 under the checking instructions. Under the field-count control, if fields erred independently, at most 5.8\% of these records would be wrong in every field. Relative to the general check, the path check changes VPA by $+3.0$ points for Qwen (95\% interval $[-1.0, 6.8]$) and $+3.3$ for Gemma ($[-1.0, 8.1]$), over 61 and 60 paired instances; both intervals include zero. Qwen's 96-by-2 configuration has nine truncated outputs, so its analysis covers only 41.9\% of records.

\begin{table}[t]
  \centering
  \small
  \setlength{\tabcolsep}{2pt}
  \renewcommand{\arraystretch}{1.12}
  \begin{tabular*}{\columnwidth}{@{\extracolsep{\fill}}l l rrrr@{}}
    \toprule
    Model & Cond. & Scor. & VPA (\%) & EM & Whole \\
    \midrule
    \multirow{5}{*}{Qwen 9B}
      & $96\times2$   & 15/24 & 62.43 & 0/24 & 67/67 \\
      & $48\times4$   & 24/24 & 64.08 & 0/24 & 114/114 \\
      & $24\times8$   & 24/24 & 71.68 & 0/24 & 82/83 \\
      & General & 66/72 & 63.40 & 0/72 & 310/311 \\
      & Path & 64/72 & 66.92 & 0/72 & 252/252 \\
    \addlinespace[4pt]
    \multirow{5}{*}{Gemma 12B}
      & $96\times2$   & 24/24 & 67.06 & 0/24 & 189/189 \\
      & $48\times4$   & 22/24 & 67.80 & 0/24 & 112/112 \\
      & $24\times8$   & 24/24 & 55.21 & 0/24 & 89/89 \\
      & General & 66/72 & 66.79 & 1/72 & 141/141 \\
      & Path & 65/72 & 71.80 & 0/72 & 82/82 \\
    \bottomrule
  \end{tabular*}
  \caption{Field-count and checking controls. $r\times f$: records per instance $\times$ fields per record. Whole: records moved whole / records with at least one misplaced field. General and Path are the checking instructions, pooled over the three configurations.}
  \label{tab:record-controls}
\end{table}

\subsection{Path Similarity and Input Position}
\label{app:record-risk}

These analyses use the 394 scorable outputs of the field-count and checking controls (21,096 records). Each record is identified by its complete field--value tuple, and every prefix containing that tuple is found, so omissions and copies can be traced as well. Only 39 records (0.18\%) appear in part, as fragments elsewhere or as some fields at the target.

Table~\ref{tab:record-risk} reports both analyses. For records absent from the entire output, the vacated target is usually occupied by another complete record. Two paths are similar when their identifier tuples differ in exactly one position. The reference replaces each occupant with a record drawn uniformly from the same instance. Input adjacency shows no consistent excess: within three input rows, Qwen has 251 pairs against 318 expected, and Gemma 156 against 152. For position, the first 20\% of input records are compared with the rest; intervals use 2,000 bootstrap draws over seeds. All 24 seeds show higher early loss for both models.

\begin{table}[t]
  \centering
  \small
  \setlength{\tabcolsep}{1.5pt}
  \renewcommand{\arraystretch}{1.12}
  \begin{tabular*}{\columnwidth}{@{\extracolsep{\fill}}l rr rr@{}}
    \toprule
    & \multicolumn{2}{c}{\textbf{Path similarity}} & \multicolumn{2}{c}{\textbf{Input position}} \\
    \cmidrule(lr){2-3}\cmidrule(l){4-5}
    & Occupied & One ID (\%) & Loss (\%) & Ratio \\
    \midrule
    Qwen& 2,122/2,340 & 86.1 / 11.8 & 60.2 / 14.1 & 4.3 [3.7, 5.0] \\
    Gemma& 787/1,616   & 91.2 / 10.6 & 30.0 / 10.3 & 2.9 [2.5, 3.4] \\
    \bottomrule
  \end{tabular*}
  \caption{Path similarity and input position. Occupied: absent records whose target is taken by another record. One ID: occupants differing from the absent record by one identifier, observed / reference, over 3,136 (Qwen 9B) and 1,850 (Gemma 12B) pairs. Loss: absence rate of the first 20\% of input records / the rest, with the ratio and 95\% interval.}
  \label{tab:record-risk}
\end{table}

\subsection{Unique Identifiers}
\label{app:record-unique}

The diagnostic uses the same new instances as the validation in \S\ref{sec:interventions}, with shuffled inputs and nested outputs. On Original, order, shipment, and package identifiers become unique within the instance; tenant identifiers are already unique, and the sender and recipient roles stay fixed. On Deep--narrow, every grouping level becomes unique. Field values, record order, and tree shape are unchanged, but identifiers and target paths differ from the baseline.

\begin{table}[t]
  \centering
  \small
  \setlength{\tabcolsep}{2pt}
  \renewcommand{\arraystretch}{1.12}
  \begin{tabular*}{\columnwidth}{@{\extracolsep{\fill}}l l rr rrr@{}}
    \toprule
    & & & & \multicolumn{3}{c}{\textbf{Record errors}} \\
    \cmidrule(l){5-7}
    Struct. & IDs & EM & VPA (\%) & Other & Outside & Omitted \\
    \midrule
    \multicolumn{7}{@{}l}{\textit{Qwen3.5-9B}} \\
    Original     & Base   & 0/24 & 80.82 & 111 & 0   & 110 \\
                 & Unique & 3/24 & 89.93 & 8   & 81  & 27 \\
    Deep & Base   & 0/24 & 60.33 & 121 & 3   & 286 \\
                 & Unique & 0/24 & 90.08 & 0   & 64  & 36 \\
    \addlinespace[4pt]
    \multicolumn{7}{@{}l}{\textit{Gemma 4 12B IT}} \\
    Original     & Base   & 3/24 & 68.49 & 5   & 333 & 25 \\
                 & Unique & 0/24 & 47.06 & 0   & 558 & 1 \\
    Deep & Base   & 0/24 & 62.95 & 86  & 147 & 175 \\
                 & Unique & 0/24 & 83.81 & 0   & 162 & 9 \\
    \addlinespace[4pt]
    \multicolumn{7}{@{}l}{\textit{Ministral 3 8B Instruct}} \\
    Original     & Base   & 0/24 & 79.62 & 91  & 59  & 75 \\
                 & Unique & 7/24 & 93.99 & 36  & 25  & 8 \\
    Deep & Base   & 0/24 & 54.08 & 197 & 214 & 118 \\
                 & Unique & 0/24 & 85.59 & 0   & 130 & 36 \\
    \bottomrule
  \end{tabular*}
  \caption{Unique-identifier diagnostic on new instances. Columns as in Table~\ref{tab:interventions}; Base rows repeat its baseline. Deep: Deep--narrow.}
  \label{tab:unique}
\end{table}

Table~\ref{tab:unique} gives the full results. Records written to other valid paths fall sharply for all three models, yet exact matches change little, because omitted and out-of-tree records remain. With unique identifiers, a record also falls outside the tree when its identifiers are combined under a wrong parent, so out-of-tree counts in this diagnostic do not measure closure errors alone.

\section{Hierarchy-Realization Diagnostic Details}
\label{app:hierarchy}

\begin{figure*}[t]
  \centering
  \begin{minipage}[t]{0.49\textwidth}
    \footnotesize
    \textbf{Shared hierarchy rules}\par\vspace{2pt}
\begin{verbatim}
root ::= ws "{" ws "\"tenants\"" colon
         tenant-map ws "}" ws
tenant-map ::= "{" ws ten
               (comma ten)* ws "}"
ten ::= "\"t" [0-9]+ "\"" colon
        "{" ws "\"orders\"" colon
        order-map ws "}"
order-map ::= "{" ws ord
              (comma ord)* ws "}"
ord ::= "\"o" [0-9]+ "\"" colon
        "{" ws "\"shipments\"" colon
        shipment-map ws "}"
shipment-map ::= "{" ws shp
                 (comma shp)* ws "}"
shp ::= "\"s" [0-9]+ "\"" colon
        "{" ws "\"packages\"" colon
        package-map ws "}"
package-map ::= "{" ws pkg
                (comma pkg)* ws "}"
addr ::= "{" ws
         "\"city\"" colon str comma
         "\"street\"" colon str comma
         "\"postal_code\"" colon str
         comma
         "\"contact_name\"" colon str
         ws "}"
str ::= "\"" [^"\\\n]* "\""
colon ::= ws ":" ws
comma ::= ws "," ws
ws ::= [ \t\n]{0,80}
\end{verbatim}
  \end{minipage}\hfill
  \begin{minipage}[t]{0.49\textwidth}
    \footnotesize
    \textbf{Original: package and role rules}\par\vspace{2pt}
\begin{verbatim}
pkg ::= "\"p" [0-9]+ "\"" colon
        "{" ws "\"addresses\"" colon
        "{" ws
        "\"sender\"" colon addr comma
        "\"recipient\"" colon addr ws "}"
        ws "}"
\end{verbatim}
    \textbf{Deep--narrow: replacement rules}\par\vspace{2pt}
\begin{verbatim}
pkg ::= "\"p" [0-9]+ "\"" colon
        "{" ws "\"addresses\"" colon
        address-map ws "}"
address-map ::= "{" ws adr
                (comma adr)* ws "}"
adr ::= "\"a" [0-9]+ "\"" colon addr
\end{verbatim}
    \textbf{T+C: replace the tenant-map rule}\par\vspace{2pt}
\begin{verbatim}
tenant-map ::= "{" ws
               ten comma ten ws "}"
\end{verbatim}
    T uses the shared rules with the package rules of its structure; T+C also replaces \texttt{tenant-map}.
  \end{minipage}
  \caption{Tree-shape grammars (EBNF). T and T+C differ only in the \texttt{tenant-map} rule.}
  \label{fig:hierarchy-grammar}
\end{figure*}

This appendix gives the setup and full results of the constrained-decoding diagnostic in \S\ref{sec:hierarchy-realization}.

\subsection{Setup and Predictions}
\label{app:hierarchy-setup}

Gemma 12B and Ministral 8B each solve 24 new Original instances and 24 new Deep--narrow instances. Each instance has 48 records under two tenants, and inputs follow the target order. Every instance is decoded under four conditions with the same prompt, input, and reference, giving 384 outputs. Decoding uses vLLM 0.23.0 \citep{kwon2023efficient} with the xgrammar 0.2.3 backend \citep{dong2025xgrammar}.

Predictions were fixed before the instances were generated. If closure errors explain the remaining failures under aligned input, the tree-shape grammar plus tenant count should give Gemma at least 22/24 on each structure with almost no records on other valid paths, the tree-shape grammar alone should give Ministral at least 22/24, and Gemma's main change under it should be whole-tenant omission. Requiring only valid JSON should not raise exact matches by more than three per cell. Frequent misplacement to other valid paths, or four or more truncations in a cell, would count against this account.

\subsection{Decoding Conditions}
\label{app:hierarchy-constraints}

\paragraph{Unconstrained (U).} No output constraint; the prompt still specifies the target structure.

\paragraph{JSON only (J).} The output must be a JSON object, with no constraint on containers or nesting.

\paragraph{Tree shape (T).} An EBNF grammar fixes the root key, the nesting of the five containers, the field names and their order, and the identifier prefixes (Figure~\ref{fig:hierarchy-grammar}). Original also fixes the sender and recipient roles. Every other entity map must be nonempty but has no fixed size.

\paragraph{Tree shape plus count (T+C).} The grammar additionally requires exactly two tenant entries, a count readable from the input.

Neither grammar contains the instance's identifiers, record bindings, or field values; the model still chooses which records go where.

\subsection{Closure Detection}
\label{app:hierarchy-tracing}

Premature closure is flagged when the root contains a key other than the top-level container. Under-closure is flagged when an entity key appears directly under another entity key without its container. The two flags can co-occur in one output. Records are traced by their complete field--value tuples as in Appendix~\ref{app:record-risk}, and a whole-tenant omission is flagged when none of a tenant's records appears anywhere in the output.

\subsection{Results}
\label{app:hierarchy-results}

\begin{table}[t]
  \centering
  \small
  \setlength{\tabcolsep}{3pt}
  \renewcommand{\arraystretch}{1.12}
  \begin{tabular*}{\columnwidth}{@{\extracolsep{\fill}}l l rrrrr@{}}
    \toprule
    Structure & Cond. & EM & PC & UC & Omit. & Outside \\
    \midrule
    \multicolumn{7}{@{}l}{\textit{Gemma 4 12B IT}} \\
    Original     & U   & 10/24 & 14 & 3 & 0  & 336 \\
                 & J   & 8/24  & 16 & 4 & 0  & 384 \\
                 & T   & 11/24 & 0  & 0 & 13 & 0 \\
                 & T+C & 24/24 & 0  & 0 & 0  & 0 \\
    Deep--narrow & U   & 1/24  & 21 & 7 & 0  & 504 \\
                 & J   & 2/24  & 20 & 9 & 0  & 480 \\
                 & T   & 1/24  & 0  & 0 & 23 & 0 \\
                 & T+C & 24/24 & 0  & 0 & 0  & 0 \\
    \addlinespace[4pt]
    \multicolumn{7}{@{}l}{\textit{Ministral 3 8B Instruct}} \\
    Original     & U   & 22/24 & 0 & 2  & 0 & 80 \\
                 & J   & 22/24 & 0 & 2  & 0 & 80 \\
                 & T   & 24/24 & 0 & 0  & 0 & 0 \\
                 & T+C & 24/24 & 0 & 0  & 0 & 0 \\
    Deep--narrow & U   & 0/24  & 0 & 23 & 0 & 630 \\
                 & J   & 0/24  & 0 & 23 & 0 & 642 \\
                 & T   & 24/24 & 0 & 0  & 0 & 0 \\
                 & T+C & 24/24 & 0 & 0  & 0 & 0 \\
    \bottomrule
  \end{tabular*}
  \caption{Constrained-decoding diagnostic. U: unconstrained; J: JSON only; T: tree shape; T+C: tree shape plus tenant count. PC, UC: outputs flagged for premature closure and under-closure. Omit.: outputs missing a whole tenant. Outside: records outside the target tree.}
  \label{tab:hierarchy-results}
\end{table}

All 384 outputs end normally, with no truncation or invalid JSON, and no output in any condition places a record on another valid path (Table~\ref{tab:hierarchy-results}). Requiring only valid JSON leaves exact matches and closure errors close to the unconstrained level. The tree-shape grammar makes Ministral correct on every instance. For Gemma, it removes both closure flags, but 13 Original and 23 Deep--narrow outputs end with one tenant's 24 records missing, at roughly half the usual length. Paired by instance, 13 of the 14 Original outputs with premature closure under U become tenant omissions under T, and the remaining one becomes correct. Adding the tenant count makes both models correct on every instance.

All prediction criteria are met. The grammar still permits wrong identifiers, parents, and values, so these exact matches are observed outcomes rather than guarantees of the grammar.

\clearpage
\vsize=\textheight
\section{Intervention Details}
\label{app:interventions}

This appendix supplies the construction, generation settings, scoring denominators, and complete condition-level results for Section~\ref{sec:interventions}. The fresh 2$\times$2 validation, local-shuffle check, large-model record-count gradient, PDB validation, and MBTA ordering check are reported separately. Reused contents and repeated conditions do not constitute additional independent instances.

\subsection{Interventions and Reporting Protocol}
\label{app:intervention-protocol}

\paragraph{The four conditions.}
\textbf{Base} uses shuffled input records and nested JSON output. \textbf{Order} sorts the same input records by the target's hierarchical traversal and retains nested output. \textbf{Path} retains shuffled input but requests a flat object whose keys are complete RFC~6901 \emph{record} pointers and whose values contain all payload fields of one record. \textbf{Both} combines sorted input with record-pointer output. Thus, Path emits one entry per record, rather than one entry per leaf field. A deterministic program assembles the nested target from the declared pointers; malformed or conflicting declarations remain subject to the representation and scoring checks in Appendix~\ref{app:scoring}.

Within a base instance, all four conditions preserve the record contents, grouping identifiers, intended parent assignments, and nested reference. Order changes only the input permutation. Path changes the requested output representation and its structural example; it neither supplies the correct instance-specific pointers nor restricts decoding through a schema or grammar. The logistics prompts retain the copying and inclusion rules. PDB prompts likewise retain all eight atom payload fields and all four native grouping fields. Unique-identifier and leaf-pointer diagnostics have different interventions and are not included in this 2$\times$2 comparison.

\begin{table*}[t]
  \centering
  \small
  \setlength{\tabcolsep}{4pt}
  \renewcommand{\arraystretch}{1.15}
  \begin{tabularx}{\textwidth}{>{\raggedright\arraybackslash}Xlrr>{\raggedright\arraybackslash}X}
    \toprule
    \textbf{Comparison} & \textbf{Records / fields} & \textbf{Base instances} & \textbf{Outputs} & \textbf{Pairing and reuse} \\
    \midrule
    Fresh 2$\times$2 & 48 / 4 & 48 & 576 & 2 structures $\times$ 24; 3 models $\times$ 4 arms. \\
    Local shuffling & 48 / 4 & 24 & 144 & Earlier Original contents; 2 models $\times$ 3 orders. \\
    Large-model gradient & 48, 64, 96 / 4 & 72 & 576 & 3 levels $\times$ 24; 2 models $\times$ 4 arms. The 48-record contents overlap fresh validation. \\
    PDB & 48--66 / 8 & 48 & 576 & 47 eligible entries plus 1 supplement; 3 models $\times$ 4 arms. \\
    MBTA & 48 / 4 & 24 & 192 & 2 paired trip-ID variants $\times$ 2 models $\times$ 2 ordering arms. \\
    \bottomrule
  \end{tabularx}
  \par\vspace{4pt}
  \begin{minipage}{\textwidth}
    \footnotesize
    \textit{Notes.} Fields counts payload fields per record. Outputs counts generated attempts selected for these comparisons, including failures. Base instances are shared across models and conditions; overlapping cohorts are not pooled into an independent-sample denominator. Unique-ID, leaf-pointer, probe, and earlier exploratory arms are excluded from these counts.
  \end{minipage}
  \caption{\textbf{Intervention cohorts and exact reporting scope.} The appendix keeps distinct validation purposes and reused contents explicit.}
  \label{tab:intervention-cohorts}
\end{table*}

\paragraph{Generation.}
All cohorts use vLLM \citep{kwon2023efficient}, temperature 0, sampling seed 42, a 16,384-token output budget, and a 1,200-second request timeout. Qwen and Gemma use disabled thinking; Ministral omits the unsupported thinking parameter. Gemma and Ministral receive the explicit JSON-only system instruction consistently across the arms of their comparisons. This instruction does not impose a decoding constraint. Fresh validation and PDB use concurrency 32; local shuffling and native-ID MBTA use 8. Large-model concurrency is 32 for Qwen 27B and 16 for Gemma 31B; local-trip MBTA uses 32. Fresh validation, the large-model gradient, and PDB use streamed responses. No result-dependent stopping, retries, seed replacement, or output repair is used. Scoring removes a Markdown fence enclosing the complete response when present, following the shared protocol.

\paragraph{Accuracy and output length.}
EM counts fully correct outputs over \emph{all} attempts. Invalid JSON and truncated responses remain EM failures and are excluded from VP, VPA, output-leaf precision, and record tracing. The three leaf metrics are macro-averages over scorable outputs; a higher conditional mean with fewer scorable outputs need not represent an improvement. Tables display both the scorable denominator and invalid/truncated counts. Output length is the rounded median number of generated tokens among normally completed responses, including completed but invalid JSON and excluding truncations.

\paragraph{Record attribution.}
Record diagnostics use the complete typed field--value tuple, unchanged field names, and the evaluator's retained duplicate declarations, as in Appendix~\ref{app:record-assignment}. \textbf{Other} counts records missing clean target placement whose first uniquely attributable complete occurrence is at another valid reference record prefix. \textbf{Outside} counts the corresponding occurrences outside the reference prefix set. \textbf{Absent} counts records with no uniquely attributable complete occurrence anywhere and no correct field at their target; surviving fragments can still exist. \textbf{Conflict} includes a complete target tuple accompanied by inconsistent declarations, or a target containing every payload-field name with at least one correct field and at least one mixed or corrupted field. Additional complete occurrences, including repeated declarations at the target, are counted separately as \textbf{Copies}. The first off-target occurrence of a record missing at its target is its movement; later off-target occurrences are copies. An off-target destination prefix is counted once per source tuple; excess complete declarations at the target are counted separately. These are descriptions of output forms, not inferred internal operations. Correct placement of all source tuples does not imply exact match, which also checks extra fields and structural declarations.

Source records with identical complete tuples, and ambiguous combinations of multiple records at one output prefix, cannot support unique movement attribution. In the scorable outputs reported here, every source tuple is unique within its instance. All reference records enter the tuple audit, with the per-row denominator $N$ shown explicitly; no residual partial-at-target category remains after mixed-value conflicts are separated. Repeated individual values in real data still preclude applying the stricter single-occurrence leaf criterion to every record. Consequently, this audit does not extend the 98.6\% strict movement statistic to PDB or MBTA.

\subsection{Fresh \texorpdfstring{2$\times$2}{2x2} Validation}
\label{app:intervention-fresh}

\paragraph{Instances and predictions.}
Each of Qwen3.5-9B, Gemma 4 12B IT, and Ministral 3 8B Instruct evaluates 24 Original logistics instances (seeds 20271001--20271024) and 24 Deep--narrow instances (20271101--20271124). These seeds were unused in the earlier exploratory runs. Every instance has 48 records and four payload fields, giving 192 target leaves. Original has branching $[2,3,2,2,2]$, with sender/recipient roles at the final level; Deep--narrow has branching $[2,2,2,2,3]$, with address IDs at the final level. These are the structures defined in Appendix~\ref{app:examples}.

The configuration and predictions were frozen before generation: Order should reduce assignment to other valid prefixes and whole-record loss, Path should reduce tree-exterior placement, and Both should attain the highest EM while exposing any residual errors. Ministral's hierarchy failures under Order were an explicit test of the combination. The archive contains five arms per model, including the Unique-ID diagnostic reported in Appendix~\ref{app:record-unique}. Only the four intervention arms enter this appendix: 192 outputs per model and 576 in total, from the same 48 base instances.

\begin{table*}[t]
  \centering
  \footnotesize
  \setlength{\tabcolsep}{3pt}
  \renewcommand{\arraystretch}{1.10}
  \begin{tabularx}{\textwidth}{Xllrrrrrrr}
    \toprule
    \textbf{Model} & \textbf{Structure} & \textbf{Arm} & \textbf{EM} & \textbf{S} & \textbf{I/T} & \textbf{VP} & \textbf{VPA} & \textbf{Prec.} & \textbf{Tokens} \\
    \midrule
    \textbf{Qwen 9B} & Original & Base & 0/24 & 24 & 0/0 & 90.45 & 80.82 & 81.02 & 3480 \\
     &  & Order & \textbf{24/24} & 24 & 0/0 & 100.00 & 100.00 & 100.00 & 3480 \\
     &  & Path & 0/24 & 20 & 0/4 & 89.17 & 80.42 & 84.70 & 3565 \\
     &  & \textbf{Both} & \textbf{24/24} & 24 & 0/0 & 100.00 & 100.00 & 100.00 & 3640 \\
    \addlinespace[5pt]
    \textbf{Qwen 9B} & Deep--narrow & Base & 0/24 & 23 & 0/1 & 75.36 & 60.33 & 41.56 & 6416 \\
     &  & Order & 23/24 & 24 & 0/0 & 100.00 & 99.48 & 99.48 & 3732 \\
     &  & Path & 0/24 & 16 & 0/8 & 86.59 & 72.14 & 72.74 & 4447 \\
     &  & \textbf{Both} & \textbf{24/24} & 24 & 0/0 & 100.00 & 100.00 & 100.00 & 4450 \\
    \addlinespace[5pt]
    \textbf{Gemma 12B} & Original & Base & 3/24 & 24 & 0/0 & 97.83 & 68.49 & 69.63 & 3535 \\
     &  & Order & 8/24 & 24 & 0/0 & 99.65 & 66.32 & 66.48 & 3541 \\
     &  & Path & 12/24 & 24 & 0/0 & 98.78 & 97.74 & 98.34 & 3988 \\
     &  & \textbf{Both} & \textbf{24/24} & 24 & 0/0 & 100.00 & 100.00 & 100.00 & 3989 \\
    \addlinespace[5pt]
    \textbf{Gemma 12B} & Deep--narrow & Base & 0/24 & 23 & 1/0 & 84.15 & 62.95 & 71.64 & 3488 \\
     &  & Order & 2/24 & 24 & 0/0 & 100.00 & 56.25 & 56.25 & 3805 \\
     &  & Path & 1/24 & 20 & 0/4 & 89.77 & 76.43 & 83.98 & 4401 \\
     &  & \textbf{Both} & \textbf{24/24} & 24 & 0/0 & 100.00 & 100.00 & 100.00 & 4858 \\
    \addlinespace[5pt]
    \textbf{Ministral 8B} & Original & Base & 0/24 & 23 & 1/0 & 93.21 & 79.62 & 80.16 & 3122 \\
     &  & Order & 17/24 & 24 & 0/0 & 100.00 & 78.12 & 78.12 & 3123 \\
     &  & Path & 0/24 & 23 & 0/1 & 94.18 & 84.31 & 86.37 & 3527 \\
     &  & \textbf{Both} & \textbf{24/24} & 24 & 0/0 & 100.00 & 100.00 & 100.00 & 3533 \\
    \addlinespace[5pt]
    \textbf{Ministral 8B} & Deep--narrow & Base & 0/24 & 24 & 0/0 & 89.76 & 54.08 & 57.26 & 3306 \\
     &  & Order & 2/24 & 24 & 0/0 & 100.00 & 51.04 & 51.04 & 3402 \\
     &  & Path & 0/24 & 23 & 0/1 & 93.93 & 79.96 & 81.55 & 4314 \\
     &  & \textbf{Both} & \textbf{24/24} & 24 & 0/0 & 100.00 & 100.00 & 100.00 & 4322 \\
    \bottomrule
  \end{tabularx}
  \par\vspace{4pt}
  \begin{minipage}{\textwidth}
    \footnotesize
    \textit{Notes.} EM includes all attempts; S: scorable outputs; I/T: invalid/truncated counts. Leaf metrics are percentages macro-averaged over S. Tokens is the rounded median over completed responses. API failures are zero. Bold EM marks full correctness.
  \end{minipage}
  \caption{\textbf{Fresh 2$\times$2 validation: accuracy, scoring coverage, and output length.} Each row attempts 24 outputs from 48-record instances.}
  \label{tab:intervention-fresh-scores}
\end{table*}

\begin{table*}[t]
  \centering
  \footnotesize
  \setlength{\tabcolsep}{3pt}
  \renewcommand{\arraystretch}{1.10}
  \begin{tabularx}{\textwidth}{Xllrrrrrr}
    \toprule
    \textbf{Model} & \textbf{Structure} & \textbf{Arm} & \textbf{N} & \textbf{Other} & \textbf{Outside} & \textbf{Absent} & \textbf{Conflict} & \textbf{Copies} \\
    \midrule
    \textbf{Qwen 9B} & Original & Base & 1152 & 111 & 0 & 110 & 0 & 107 \\
     &  & Order & 1152 & 0 & 0 & 0 & 0 & 0 \\
     &  & Path & 960 & 84 & 0 & 104 & 0 & 60 \\
     &  & \textbf{Both} & 1152 & 0 & 0 & 0 & 0 & 0 \\
    \addlinespace[5pt]
    \textbf{Qwen 9B} & Deep--narrow & Base & 1104 & 121 & 3 & 286 & 28 & 1085 \\
     &  & Order & 1152 & 6 & 0 & 0 & 0 & 0 \\
     &  & Path & 768 & 107 & 0 & 103 & 4 & 93 \\
     &  & \textbf{Both} & 1152 & 0 & 0 & 0 & 0 & 0 \\
    \addlinespace[5pt]
    \textbf{Gemma 12B} & Original & Base & 1152 & 5 & 333 & 25 & 0 & 7 \\
     &  & Order & 1152 & 0 & \textbf{384} & 4 & 0 & 0 \\
     &  & Path & 1152 & 10 & 0 & 14 & 2 & 7 \\
     &  & \textbf{Both} & 1152 & 0 & 0 & 0 & 0 & 0 \\
    \addlinespace[5pt]
    \textbf{Gemma 12B} & Deep--narrow & Base & 1104 & 86 & 147 & 175 & 1 & 41 \\
     &  & Order & 1152 & 0 & \textbf{504} & 0 & 0 & 0 \\
     &  & Path & 960 & 80 & 23 & 98 & 26 & 13 \\
     &  & \textbf{Both} & 1152 & 0 & 0 & 0 & 0 & 0 \\
    \addlinespace[5pt]
    \textbf{Ministral 8B} & Original & Base & 1104 & 91 & 59 & 75 & 0 & 68 \\
     &  & Order & 1152 & 0 & \textbf{252} & 0 & 0 & 0 \\
     &  & Path & 1104 & 100 & 1 & 64 & 9 & 37 \\
     &  & \textbf{Both} & 1152 & 0 & 0 & 0 & 0 & 0 \\
    \addlinespace[5pt]
    \textbf{Ministral 8B} & Deep--narrow & Base & 1152 & 197 & 214 & 118 & 0 & 51 \\
     &  & Order & 1152 & 0 & \textbf{564} & 0 & 0 & 0 \\
     &  & Path & 1104 & 149 & 1 & 68 & 4 & 44 \\
     &  & \textbf{Both} & 1152 & 0 & 0 & 0 & 0 & 0 \\
    \bottomrule
  \end{tabularx}
  \par\vspace{4pt}
  \begin{minipage}{\textwidth}
    \footnotesize
    \textit{Notes.} N covers scorable reference records. Other/Outside/Absent/Conflict are mutually exclusive; Copies is separate. Source tuples are unique within each instance. Definitions and ambiguous-occurrence handling follow Appendix~\ref{app:intervention-protocol}.  Maximum N is 1,152 per row; invalid or truncated outputs reduce this denominator. Bold Outside counts highlight residual hierarchy errors under Order.
  \end{minipage}
  \caption{\textbf{Fresh 2$\times$2 validation: complete record diagnostics.} The three record-error columns in the main-text table are supplemented by target conflicts and extra copies.}
  \label{tab:intervention-fresh-records}
\end{table*}

\paragraph{Complementarity and remaining failures.}
Tables~\ref{tab:intervention-fresh-scores} and~\ref{tab:intervention-fresh-records} reproduce the main-text comparison with separate denominators, conflicts, copies, and lengths. Under Order, Qwen attains \textbf{47/48} exact matches; its one remaining Deep--narrow failure places six records at other valid prefixes. Gemma and Ministral also lose almost all assignment and omission errors, but their tree-exterior counts rise to 384/504 and 252/564 on Original/Deep--narrow. Removing input disorder therefore does not by itself remove these models' synthetic hierarchy failures.

Under Path, tree-exterior placement is zero or nearly zero in most cells, but wrong valid-prefix placement and missing records remain. On Original, Gemma improves from 3/24 to 12/24 EM; on Deep--narrow, it improves only from 0/24 to 1/24. Neither Qwen nor Ministral obtains an exact match under Path on either structure. \textbf{Both reaches 24/24 in every model--structure cell}, with no record errors, conflicts, or copies.

Coverage matters particularly for Path. Base has one Qwen Deep--narrow truncation and one invalid output each for Gemma Deep--narrow and Ministral Original. Order and Both have no unscorable outputs. Path has 4/8 Qwen truncations on Original/Deep--narrow, four Gemma Deep--narrow truncations, and one Ministral truncation on each structure. Its smaller record-error counts therefore cover fewer answers in those cells. They must not be read as counts of the errors that the truncated outputs would have contained.

Even when generation completes, the representation changes length. Original Order-to-Both medians rise from 3,480 to 3,640 tokens for Qwen, 3,541 to 3,989 for Gemma, and 3,123 to 3,533 for Ministral. The corresponding Deep--narrow increases are 3,732 to 4,450, 3,805 to 4,858, and 3,402 to 4,322. Repeating a complete record path has a measurable length cost; these medians do not include the responses that reached the token limit.

\subsection{Local Shuffling on Reused Original Instances}
\label{app:intervention-local}

This check reuses the earlier 24 Original instances, with seeds 20261002--20261025. Qwen 9B and Gemma 12B each generate three outputs per instance. The records, nested reference, copying rules, structural example, and output representation remain fixed. \textbf{Shuffled} preserves the original random input permutation; \textbf{Order} follows the target traversal. \textbf{Local} starts from Order, partitions the 48 records into six contiguous blocks of eight target ranks, and deterministically shuffles within each block, preserving block membership and block order. The block-specific random seed is $\mathrm{baseSeed}\mathbin{\mathrm{XOR}}(i+421)$ for zero-based block index $i$; an unchanged permutation is rotated by one position.

The archived rank statistics confirm that Local lies between the endpoints: mean inversion counts are 566.33, 82.96, and 0 for Shuffled, Local, and Order; mean Kendall $\tau$ is $-0.0041$, $0.8529$, and $1$. Mean switches between consecutive package-parent prefixes are 45.96, 41.38, and 23 in the same order. These are observable input-order statistics, rather than direct measurements of an internal path-switching cost. Local also preserves contiguous path neighborhoods, so the comparison is not an isolated scalar manipulation of inversion count.

\begin{table*}[t]
  \centering
  \footnotesize
  \setlength{\tabcolsep}{3pt}
  \renewcommand{\arraystretch}{1.10}
  \begin{tabularx}{\textwidth}{Xlrrrrrrrrr}
    \toprule
    \textbf{Model} & \textbf{Order} & \textbf{EM} & \textbf{VP} & \textbf{VPA} & \textbf{Prec.} & \textbf{Other} & \textbf{Outside} & \textbf{Absent} & \textbf{Copies} & \textbf{Tokens} \\
    \midrule
    \textbf{Qwen 9B} & Shuffled & 0/24 & 92.27 & 83.85 & 83.85 & 97 & 0 & 89 & 89 & 3482 \\
     & Local & 10/24 & 98.18 & 96.79 & 97.14 & 16 & 0 & 21 & 17 & 3481 \\
     & \textbf{Order} & \textbf{24/24} & 100.00 & 100.00 & 100.00 & 0 & 0 & 0 & 0 & 3481 \\
    \addlinespace[5pt]
    \textbf{Gemma 12B} & Shuffled & 2/24 & 97.31 & 55.73 & 56.98 & 4 & 475 & 31 & 4 & 3484 \\
     & Local & 3/24 & 99.91 & 56.16 & 56.21 & 0 & 504 & 1 & 0 & 3540 \\
     & \textbf{Order} & 7/24 & 100.00 & 64.58 & 64.58 & 0 & 408 & 0 & 0 & 3541 \\
    \bottomrule
  \end{tabularx}
  \par\vspace{4pt}
  \begin{minipage}{\textwidth}
    \footnotesize
    \textit{Notes.} All rows have 24 scorable outputs and 1,152 reference records. There are no invalid outputs, truncations, API failures, target conflicts, or remaining partial records. Accuracy columns are percentages; record columns are counts. Tokens follows the same median convention as the fresh validation. These are reused Original contents, not new independent instances.
  \end{minipage}
  \caption{\textbf{Local-shuffle check on reused Original instances.} Local permutes records within contiguous target-rank blocks of eight.}
  \label{tab:intervention-local-shuffle}
\end{table*}

Qwen EM progresses \textbf{0/24 $\rightarrow$ 10/24 $\rightarrow$ 24/24}, while other-valid records decline from 97 to 16 to 0 and absent records from 89 to 21 to 0. Gemma progresses only 2/24 $\rightarrow$ 3/24 $\rightarrow$ 7/24; tree-exterior counts remain large at 475, 504, and 408. All 144 outputs are scorable and normally completed. These observations support the ordering interpretation for Qwen while retaining the hierarchy limitation for Gemma; they are a reused-content check, separate from the fresh-instance validation.

\subsection{Large Models and the Record-Count Gradient}
\label{app:intervention-large}

Qwen3.5-27B and Gemma 4 31B IT evaluate Original logistics instances with 48, 64, and 96 records, respectively 192, 256, and 384 payload leaves. Only orders per tenant changes: 3, 4, and 6. There remain two tenants, two shipments per order, two packages per shipment, and two address roles per package. Each level has 24 base instances crossed with the four conditions, yielding 288 outputs per model and 576 in total.

The 48-record level uses seeds 20271001--20271024 and has prompts identical to the corresponding small-model fresh validation. The 64-record and 96-record levels use seeds 20271401--20271424 and 20271501--20271524. Thus, the difficulty levels do not lengthen the same individual source instance. The small-model 48-record comparison reuses its earlier saved outputs; model-size comparisons are across batches and servers under the same protocol, rather than concurrent generation with all settings held identical. Large models use two-way tensor parallelism and a 32,768-token context limit.

Before generation, the configuration predicted decreasing baseline EM as record count increased, recurrence of at least one error type, and the highest EM under Both. All 576 outputs are scorable, with no truncations, invalid JSON, or API failures. Tables~\ref{tab:intervention-large-scores} and~\ref{tab:intervention-large-records} report every level rather than pooling away changes in difficulty.

\begin{table*}[t]
  \centering
  \footnotesize
  \setlength{\tabcolsep}{3pt}
  \renewcommand{\arraystretch}{1.10}
  \begin{tabularx}{\textwidth}{Xllrrrrrrr}
    \toprule
    \textbf{Model} & \textbf{Records} & \textbf{Arm} & \textbf{EM} & \textbf{S} & \textbf{I/T} & \textbf{VP} & \textbf{VPA} & \textbf{Prec.} & \textbf{Tokens} \\
    \midrule
    \textbf{Qwen 27B} & 48 & Base & 23/24 & 24 & 0/0 & 99.91 & 99.91 & 99.91 & 3480 \\
     &  & Order & \textbf{24/24} & 24 & 0/0 & 100.00 & 100.00 & 100.00 & 3480 \\
     &  & Path & 17/24 & 24 & 0/0 & 99.39 & 99.22 & 99.74 & 3638 \\
     &  & \textbf{Both} & \textbf{24/24} & 24 & 0/0 & 100.00 & 100.00 & 100.00 & 3640 \\
    \addlinespace[5pt]
    \textbf{Qwen 27B} & 64 & Base & 18/24 & 24 & 0/0 & 99.35 & 99.15 & 99.74 & 4635 \\
     &  & Order & \textbf{24/24} & 24 & 0/0 & 100.00 & 100.00 & 100.00 & 4635 \\
     &  & Path & 21/24 & 24 & 0/0 & 99.67 & 99.48 & 99.80 & 4865 \\
     &  & \textbf{Both} & \textbf{24/24} & 24 & 0/0 & 100.00 & 100.00 & 100.00 & 4865 \\
    \addlinespace[5pt]
    \textbf{Qwen 27B} & 96 & Base & 19/24 & 24 & 0/0 & 99.78 & 99.52 & 99.70 & 6970 \\
     &  & Order & \textbf{24/24} & 24 & 0/0 & 100.00 & 100.00 & 100.00 & 6970 \\
     &  & Path & 15/24 & 24 & 0/0 & 99.39 & 99.26 & 99.74 & 7340 \\
     &  & \textbf{Both} & \textbf{24/24} & 24 & 0/0 & 100.00 & 100.00 & 100.00 & 7340 \\
    \addlinespace[5pt]
    \textbf{Gemma 31B} & 48 & Base & 19/24 & 24 & 0/0 & 99.57 & 99.48 & 99.91 & 3540 \\
     &  & Order & \textbf{24/24} & 24 & 0/0 & 100.00 & 100.00 & 100.00 & 3541 \\
     &  & Path & 23/24 & 24 & 0/0 & 99.91 & 99.91 & 100.00 & 3988 \\
     &  & \textbf{Both} & \textbf{24/24} & 24 & 0/0 & 100.00 & 100.00 & 100.00 & 3989 \\
    \addlinespace[5pt]
    \textbf{Gemma 31B} & 64 & Base & 19/24 & 24 & 0/0 & 99.67 & 99.67 & 100.00 & 4717 \\
     &  & Order & \textbf{24/24} & 24 & 0/0 & 100.00 & 100.00 & 100.00 & 4717 \\
     &  & Path & 20/24 & 24 & 0/0 & 99.74 & 99.74 & 100.00 & 5331 \\
     &  & \textbf{Both} & \textbf{24/24} & 24 & 0/0 & 100.00 & 100.00 & 100.00 & 5331 \\
    \addlinespace[5pt]
    \textbf{Gemma 31B} & 96 & Base & 18/24 & 24 & 0/0 & 99.70 & 99.70 & 100.00 & 7096 \\
     &  & Order & \textbf{24/24} & 24 & 0/0 & 100.00 & 100.00 & 100.00 & 7096 \\
     &  & Path & 19/24 & 24 & 0/0 & 99.78 & 99.78 & 100.00 & 8042 \\
     &  & \textbf{Both} & \textbf{24/24} & 24 & 0/0 & 100.00 & 100.00 & 100.00 & 8042 \\
    \bottomrule
  \end{tabularx}
  \par\vspace{4pt}
  \begin{minipage}{\textwidth}
    \footnotesize
    \textit{Notes.} EM includes all attempts; S: scorable outputs; I/T: invalid/truncated counts. Leaf metrics are percentages macro-averaged over S. Tokens is the rounded median over completed responses. API failures are zero. Bold EM marks full correctness. Each row attempts 24 outputs.
  \end{minipage}
  \caption{\textbf{Large-model record-count gradient: all four conditions.} Each record contains four payload fields.}
  \label{tab:intervention-large-scores}
\end{table*}

\begin{table*}[t]
  \centering
  \footnotesize
  \setlength{\tabcolsep}{3pt}
  \renewcommand{\arraystretch}{1.10}
  \begin{tabularx}{\textwidth}{Xllrrrrrr}
    \toprule
    \textbf{Model} & \textbf{Records} & \textbf{Arm} & \textbf{N} & \textbf{Other} & \textbf{Outside} & \textbf{Absent} & \textbf{Conflict} & \textbf{Copies} \\
    \midrule
    \textbf{Qwen 27B} & 48 & Base & 1152 & 0 & 0 & 1 & 0 & 1 \\
     &  & Path & 1152 & 2 & 0 & 7 & 0 & 1 \\
    \addlinespace[5pt]
    \textbf{Qwen 27B} & 64 & Base & 1536 & 3 & 0 & 10 & 0 & 1 \\
     &  & Path & 1536 & 3 & 0 & 5 & 0 & 0 \\
    \addlinespace[5pt]
    \textbf{Qwen 27B} & 96 & Base & 2304 & 6 & 0 & 5 & 0 & 1 \\
     &  & Path & 2304 & 3 & 0 & 14 & 0 & 3 \\
    \addlinespace[5pt]
    \textbf{Gemma 31B} & 48 & Base & 1152 & 1 & 0 & 5 & 0 & 0 \\
     &  & Path & 1152 & 0 & 0 & 1 & 0 & 0 \\
    \addlinespace[5pt]
    \textbf{Gemma 31B} & 64 & Base & 1536 & 0 & 0 & 5 & 0 & 0 \\
     &  & Path & 1536 & 0 & 0 & 4 & 0 & 0 \\
    \addlinespace[5pt]
    \textbf{Gemma 31B} & 96 & Base & 2304 & 0 & 0 & 7 & 0 & 0 \\
     &  & Path & 2304 & 0 & 0 & 5 & 0 & 0 \\
    \bottomrule
  \end{tabularx}
  \par\vspace{4pt}
  \begin{minipage}{\textwidth}
    \footnotesize
    \textit{Notes.} N covers scorable reference records. Other/Outside/Absent/Conflict are mutually exclusive; Copies is separate. Source tuples are unique within each instance. Definitions and ambiguous-occurrence handling follow Appendix~\ref{app:intervention-protocol}.  All outputs are scorable. Order and Both have zero counts in every error and copy column at all three levels; their N is the same as the displayed rows.
  \end{minipage}
  \caption{\textbf{Large-model record diagnostics by difficulty level.} Base and Path are displayed; the two uniformly correct arms are specified in the notes.}
  \label{tab:intervention-large-records}
\end{table*}

Order and Both each attain \textbf{72/72 per model}. Base has no tree-exterior records at any level: Qwen has 0/3/6 other-valid records and 1/10/5 absent records at 48/64/96; Gemma has 1/0/0 other-valid records and 5/5/7 absent records. The residual failures are assignment or completeness failures with little hierarchy error for Path to remove. Path changes Qwen EM from 23/18/19 to 17/21/15, and Gemma from 19/19/18 to 23/20/19. The strict monotonic-difficulty prediction is \emph{not} supported for Qwen, whose Base EM rises from 18 to 19 at the last level. The results support the narrower observation that high leaf accuracy can coexist with a few whole-record failures, and that Path alone has no consistent advantage when hierarchy failures are absent.

\subsection{PDB Atom-Record Validation}
\label{app:intervention-pdb}

\paragraph{Source selection and sample scale.}
The source is protein atom rows from archived Protein Data Bank (PDB) mmCIF files \citep{berman2000pdb}. The candidate policy fixes seed 20261008, entries initially released in 2024, at most 100,000 deposited atoms per entry, and 24 X-ray diffraction plus 24 solution-NMR entries. Previously explored PDB IDs are excluded by identifier. Eligible protein entities are \texttt{polypeptide(L)} or \texttt{polypeptide(D)}. Missing grouping fields are reported before task construction; protein HETATM records are not automatically removed.

Models and chains are hash-ranked, taking at most two models and two chains per model. Complete residue groups are hash-ranked within the selected model--chain branches and sampled round-robin until at least 48 atom rows have been collected. Residues are never split, so the actual tasks contain \textbf{48--66 records}, rather than exactly 48. Each record supplies eight values, yielding 384--528 target leaves per task and 2,574 records across the final 48 entries. Input permutation is deterministic from the seed, source entry, and original atom ID.

\paragraph{The 47+1 denominator.}
The original frozen 48-candidate batch excludes \texttt{8TLX}: its fixed sample contains alternate conformations and colliding native four-level paths. The entire sample is excluded; no atom is deleted, conformer selected, or key renamed to make it eligible. This leaves 47 entries. A \emph{post-hoc} one-entry supplement, \texttt{8PO8}, is the first eligible remaining X-ray entry under the original hash ranking, selected using source eligibility before its model calls. The combined results therefore consist of \textbf{47 original eligible entries plus one supplemental entry}, with the original exclusion retained. They are not an unchanged prespecified 48-entry sample. Earlier 12-entry feasibility results and running snapshots are excluded.

The final X-ray IDs are \texttt{7G7H}, \texttt{7H4B}, \texttt{7HM5}, \texttt{8CHT}, \texttt{8HLR}, \texttt{8IPF}, \texttt{8J4J}, \texttt{8JM4}, \texttt{8JNK}, \texttt{8K3K}, \texttt{8OHJ}, \texttt{8PO8}, \texttt{8QUJ}, \texttt{8RDS}, \texttt{8RVM}, \texttt{8S2W}, \texttt{8UUB}, \texttt{8V7W}, \texttt{8VDT}, \texttt{8W31}, \texttt{8WVR}, \texttt{8YJO}, \texttt{8Z4Q}, and \texttt{9G91}. The NMR IDs are \texttt{8EO9}, \texttt{8JZM}, \texttt{8K81}, \texttt{8OSQ}, \texttt{8OZZ}, \texttt{8PI0}, \texttt{8QTQ}, \texttt{8R89}, \texttt{8T3H}, \texttt{8UM1}, \texttt{8UMA}, \texttt{8UWF}, \texttt{8V0H}, \texttt{8X8T}, \texttt{9BB4}, \texttt{9BHN}, \texttt{9BVU}, \texttt{9DE2}, \texttt{9F1G}, \texttt{9FJN}, \texttt{9FLA}, \texttt{9G2N}, \texttt{9G5P}, and \texttt{9GEC}.

\paragraph{Conversion task.}
Each input row retains the four native grouping fields \texttt{pdbx\_PDB\_model\_num}, \texttt{label\_asym\_id}, \texttt{label\_seq\_id}, and \texttt{label\_atom\_id}. They determine the model/chain/residue/atom keys in the target. The eight payload fields, in output order, are \texttt{id}, \texttt{type\_symbol}, \texttt{label\_comp\_id}, \texttt{Cartn\_x}, \texttt{Cartn\_y}, \texttt{Cartn\_z}, \texttt{occupancy}, and \texttt{B\_iso\_or\_equiv}. All values remain strings with their original precision. In particular, \texttt{id} is copied as payload and must not replace \texttt{label\_atom\_id} as the atom key.

Nested output uses the containers \texttt{models}, \texttt{chains}, \texttt{residues}, and \texttt{atoms}; Path uses the complete record pointer to the atom payload. Serialization follows numeric order for canonical nonnegative integer keys below $2^{32}-1$, then lexicographic order for other keys, matching the target serializer. The model performs a specified JSON transformation on supplied records, with no retrieval, biological inference, arithmetic, or coordinate conversion. Figure~\ref{fig:intervention-pdb-example} shows two unmodified source rows and both target representations.

\begin{figure*}[t]
  \centering
  \begin{minipage}[t]{0.48\textwidth}
    \textbf{Source rows from PDB entry 7G7H}
    \begin{JSONExample}[fontsize=\scriptsize]
{"atom_records": [
  {"pdbx_PDB_model_num": "1",
   "label_asym_id": "A", "label_seq_id": "362",
   "label_atom_id": "CB",
   "id": "2692", "type_symbol": "C",
   "label_comp_id": "LEU",
   "Cartn_x": "5.464", "Cartn_y": "16.334",
   "Cartn_z": "49.222", "occupancy": "1.00",
   "B_iso_or_equiv": "27.40"},
  {"pdbx_PDB_model_num": "1",
   "label_asym_id": "A", "label_seq_id": "392",
   "label_atom_id": "CB",
   "id": "2925", "type_symbol": "C",
   "label_comp_id": "HIS",
   "Cartn_x": "13.832", "Cartn_y": "16.216",
   "Cartn_z": "58.512", "occupancy": "1.00",
   "B_iso_or_equiv": "29.65"}
]}
    \end{JSONExample}
    \textbf{Record-pointer reference (Path / Both)}
    \begin{JSONExample}[fontsize=\scriptsize]
{
  "/models/1/chains/A/residues/362/atoms/CB": {
    "id": "2692", "type_symbol": "C",
    "label_comp_id": "LEU",
    "Cartn_x": "5.464", "Cartn_y": "16.334",
    "Cartn_z": "49.222", "occupancy": "1.00",
    "B_iso_or_equiv": "27.40"
  },
  "/models/1/chains/A/residues/392/atoms/CB": {
    "id": "2925", "type_symbol": "C",
    "label_comp_id": "HIS",
    "Cartn_x": "13.832", "Cartn_y": "16.216",
    "Cartn_z": "58.512", "occupancy": "1.00",
    "B_iso_or_equiv": "29.65"
  }
}
    \end{JSONExample}
  \end{minipage}\hfill
  \begin{minipage}[t]{0.48\textwidth}
    \textbf{Nested reference (Base / Order)}
    \begin{JSONExample}[fontsize=\scriptsize]
{
  "models": {"1": {
    "chains": {"A": {
      "residues": {
        "362": {"atoms": {
          "CB": {
            "id": "2692",
            "type_symbol": "C",
            "label_comp_id": "LEU",
            "Cartn_x": "5.464",
            "Cartn_y": "16.334",
            "Cartn_z": "49.222",
            "occupancy": "1.00",
            "B_iso_or_equiv": "27.40"
          }
        }},
        "392": {"atoms": {
          "CB": {
            "id": "2925",
            "type_symbol": "C",
            "label_comp_id": "HIS",
            "Cartn_x": "13.832",
            "Cartn_y": "16.216",
            "Cartn_z": "58.512",
            "occupancy": "1.00",
            "B_iso_or_equiv": "29.65"
          }
        }}
      }
    }}
  }}
}
    \end{JSONExample}
    \small
    The repeated atom name \texttt{CB} remains local to its residue. The original atom IDs \texttt{2692} and \texttt{2925} are payload values; they do not replace the atom-name keys. All eight payload strings, including numeric spelling, are copied without modification.
  \end{minipage}
  \caption{\textbf{PDB construction sample with both output representations.} Two original rows from source positions 1 and 18 of the 51-record \texttt{7G7H} task are shown as a complete two-record illustration. The evaluated task contains all 51 sampled rows; the example is not an additional evaluation instance. Container braces are compacted for display only.}
  \label{fig:intervention-pdb-example}
\end{figure*}

\begin{table*}[t]
  \centering
  \footnotesize
  \setlength{\tabcolsep}{3pt}
  \renewcommand{\arraystretch}{1.10}
  \begin{tabularx}{\textwidth}{Xlrrrrrrr}
    \toprule
    \textbf{Model} & \textbf{Arm} & \textbf{EM} & \textbf{S} & \textbf{I/T} & \textbf{VP} & \textbf{VPA} & \textbf{Prec.} & \textbf{Tokens} \\
    \midrule
    \textbf{Qwen 9B} & Base & 18/48 & 46 & 0/2 & 98.22 & 96.52 & 92.68 & 6216 \\
     & Order & \textbf{48/48} & 48 & 0/0 & 100.00 & 100.00 & 100.00 & 5986 \\
     & Path & 11/48 & 40 & 0/8 & 98.81 & 97.36 & 96.95 & 6644 \\
     & \textbf{Both} & \textbf{48/48} & 48 & 0/0 & 100.00 & 100.00 & 100.00 & 6620 \\
    \addlinespace[5pt]
    \textbf{Gemma 12B} & Base & 22/48 & 47 & 1/0 & 99.02 & 97.68 & 98.44 & 6482 \\
     & Order & \textbf{48/48} & 48 & 0/0 & 100.00 & 100.00 & 100.00 & 6452 \\
     & Path & 17/48 & 48 & 0/0 & 98.10 & 95.51 & 97.16 & 7062 \\
     & \textbf{Both} & \textbf{48/48} & 48 & 0/0 & 100.00 & 100.00 & 100.00 & 7168 \\
    \addlinespace[5pt]
    \textbf{Ministral 8B} & Base & 7/48 & 48 & 0/0 & 99.90 & 21.97 & 20.69 & 5702 \\
     & Order & 45/48 & 47 & 1/0 & 100.00 & 97.83 & 97.83 & 5547 \\
     & Path & 33/48 & 48 & 0/0 & 99.49 & 95.78 & 96.03 & 6138 \\
     & \textbf{Both} & \textbf{48/48} & 48 & 0/0 & 100.00 & 100.00 & 100.00 & 6188 \\
    \bottomrule
  \end{tabularx}
  \par\vspace{4pt}
  \begin{minipage}{\textwidth}
    \footnotesize
    \textit{Notes.} EM includes all attempts; S: scorable outputs; I/T: invalid/truncated counts. Leaf metrics are percentages macro-averaged over S. Tokens is the rounded median over completed responses. API failures are zero. Bold EM marks full correctness. Each row attempts 48 outputs: 47 original eligible entries and the one-entry supplement.
  \end{minipage}
  \caption{\textbf{PDB validation: all model--condition results.} Each source entry supplies 48--66 atom records.}
  \label{tab:intervention-pdb-scores}
\end{table*}

\begin{table*}[t]
  \centering
  \footnotesize
  \setlength{\tabcolsep}{3pt}
  \renewcommand{\arraystretch}{1.10}
  \begin{tabularx}{\textwidth}{Xlrrrrrr}
    \toprule
    \textbf{Model} & \textbf{Arm} & \textbf{N} & \textbf{Other} & \textbf{Outside} & \textbf{Absent} & \textbf{Conflict} & \textbf{Copies} \\
    \midrule
    \textbf{Qwen 9B} & Base & 2467 & 8 & 14 & 22 & 56 & 221 \\
     & Order & 2574 & 0 & 0 & 0 & 0 & 0 \\
     & Path & 2139 & 14 & 10 & 25 & 17 & 40 \\
     & \textbf{Both} & 2574 & 0 & 0 & 0 & 0 & 0 \\
    \addlinespace[5pt]
    \textbf{Gemma 12B} & Base & 2522 & 0 & 29 & 26 & 2 & 7 \\
     & Order & 2574 & 0 & 0 & 0 & 0 & 0 \\
     & Path & 2574 & 1 & 77 & 49 & 1 & 2 \\
     & \textbf{Both} & 2574 & 0 & 0 & 0 & 0 & 0 \\
    \addlinespace[5pt]
    \textbf{Ministral 8B} & Base & 2574 & 31 & \textbf{1992} & 8 & 1 & 93 \\
     & Order & 2524 & 0 & 52 & 0 & 0 & 0 \\
     & Path & 2574 & 1 & 107 & 11 & 3 & 6 \\
     & \textbf{Both} & 2574 & 0 & 0 & 0 & 0 & 0 \\
    \bottomrule
  \end{tabularx}
  \par\vspace{4pt}
  \begin{minipage}{\textwidth}
    \footnotesize
    \textit{Notes.} N covers scorable reference records. Other/Outside/Absent/Conflict are mutually exclusive; Copies is separate. Source tuples are unique within each instance. Definitions and ambiguous-occurrence handling follow Appendix~\ref{app:intervention-protocol}.  Maximum N is 2,574 per row. Complete eight-field tuples are traceable despite repeated individual atom properties. The bold count is Ministral's Base tree-exterior records.
  \end{minipage}
  \caption{\textbf{PDB record diagnostics.} N explicitly records the denominator lost through invalid or truncated outputs.}
  \label{tab:intervention-pdb-records}
\end{table*}

\paragraph{Results and attribution limits.}
All three models evaluate the same entries under four conditions: 576 outputs, rather than 576 independent source samples. Under Base, Qwen and Gemma attain 18/48 and 22/48 EM, with comparatively few other-valid, tree-exterior, and absent records. Path lowers these counts of exact matches to 11/48 and 17/48. For Qwen, the conditional VPA increase from 96.52\% to 97.36\% also excludes eight truncated outputs instead of two. Order attains \textbf{48/48 for both models}.

Ministral Base has VP 99.90\% but VPA only 21.97\%, with \textbf{1,992 tree-exterior records}. Saved outputs repeatedly use incorrect innermost atom keys despite preserving the atom payloads. Path reduces the exterior count to 107 and raises EM from 7/48 to 33/48. Order also reduces it substantially, to 52, reaching 45/48; its three failures comprise one invalid JSON response and two scorable placement failures. This qualifies the synthetic separation: on these real records, hierarchy placement is also sensitive to input order. Both attains \textbf{48/48 for all three models}; its extra EM gain over Order is confined to Ministral in this sample.

The eight-field source tuples are unique within each task, enabling the tuple audit in Table~\ref{tab:intervention-pdb-records}. This does not make every individual atom property unique. Nor are different entries guaranteed to be independent. The original candidate archive includes a sequence-similarity audit using Biopython 1.86 global alignment, at least 80\% coverage in both sequences, 90\% identity clustering, and 30\%/50\% sensitivity thresholds. It retains all candidates rather than selecting on similarity or model errors; \texttt{9GEC}, \texttt{9G2N}, and \texttt{9G5P} have close sequence relationships. This audit covers the original candidates, not a recomputed combined cohort with \texttt{8PO8}. The validation does not estimate performance over the PDB population.

\subsection{MBTA Ordering and Local Trip Identifiers}
\label{app:intervention-mbta}

\paragraph{Real stop records.}
The source is the MBTA GTFS snapshot \citep{mbtaGTFS} ``Fall 2026, version D'' dated 2026-10-02, using service date 2026-10-14 and bus routes (\texttt{route\_type=3}). Programs join routes, trips, stops, and stop times before task generation. Sampling uses seed \texttt{gtfs-mbta-v1:20261005}; the main set has 24 source groups, with routes disjoint across groups and the separate one-group probe. Only trips active on the service date and routes with both directions and enough eligible trips are retained. Special or multi-route trips, missing directions, fewer than six stops, repeated stops, invalid times, and unsafe stop text are excluded. Sampling is not filtered by value uniqueness or model outcomes.

Each task contains two routes, two directions per route, two trips per direction, and one random window of six consecutive real stops per trip: 48 stop records and 192 payload leaves. The grouping fields are \texttt{route\_id}, \texttt{direction\_id}, \texttt{trip\_id}, and \texttt{stop\_sequence}; the target containers are \texttt{routes}, \texttt{directions}, \texttt{trips}, and \texttt{stops}. Payloads contain \texttt{arrival\_time}, \texttt{stop\_id}, \texttt{stop\_name}, and \texttt{municipality}. Strings and native stop-sequence values are preserved, including windows that do not start at stop 1. Models only transform supplied stop records.

\paragraph{The local-ID variant.}
Within each route--direction pair, trips are sorted by their original IDs and relabeled \texttt{t1}, \texttt{t2}. These labels recur under different routes and directions. The native-ID and local-ID tasks preserve source payloads, stop keys, parent relationships, target traversal, and paired input permutations; only trip labels and their reference paths change. Original trip IDs remain in provenance and do not enter local-ID prompts. Relabeling is a path-distinguishability diagnostic. In contrast, the Order intervention within either variant preserves every identifier and intended target position.

Qwen 9B and Gemma 12B each evaluate Base and Order on both variants: 192 outputs from 24 reused source groups. Native-ID and local-ID results come from separate batches with concurrency 8 and 32, respectively, so the identifier comparison is not a concurrent experiment holding every execution setting fixed. Archived Unique-ID and leaf-pointer arms, the one-group probe, and the separate same-pattern variant are outside this ordering comparison; leaf-pointer output is not the record-pointer intervention used above.

\begin{table*}[t]
  \centering
  \footnotesize
  \setlength{\tabcolsep}{3pt}
  \renewcommand{\arraystretch}{1.10}
  \begin{tabularx}{\textwidth}{Xllrrrrrrr}
    \toprule
    \textbf{Trip IDs} & \textbf{Model} & \textbf{Arm} & \textbf{EM} & \textbf{S} & \textbf{I/T} & \textbf{VP} & \textbf{VPA} & \textbf{Prec.} & \textbf{Tokens} \\
    \midrule
    \textbf{Native} & Qwen 9B & Base & 8/24 & 23 & 0/1 & 99.55 & 98.53 & 92.90 & 3466 \\
     &  & \textbf{Order} & \textbf{24/24} & 24 & 0/0 & 100.00 & 100.00 & 100.00 & 3346 \\
    \addlinespace[5pt]
     & Gemma 12B & Base & \textbf{24/24} & 24 & 0/0 & 100.00 & 100.00 & 100.00 & 3443 \\
     &  & \textbf{Order} & \textbf{24/24} & 24 & 0/0 & 100.00 & 100.00 & 100.00 & 3443 \\
    \addlinespace[5pt]
    \textbf{Local trip} & Qwen 9B & Base & 3/24 & 24 & 0/0 & 98.31 & 95.42 & 86.02 & 3432 \\
     &  & \textbf{Order} & \textbf{24/24} & 24 & 0/0 & 100.00 & 100.00 & 100.00 & 3298 \\
    \addlinespace[5pt]
     & Gemma 12B & Base & 17/24 & 24 & 0/0 & 99.70 & 99.70 & 98.14 & 3396 \\
     &  & \textbf{Order} & \textbf{24/24} & 24 & 0/0 & 100.00 & 100.00 & 100.00 & 3395 \\
    \bottomrule
  \end{tabularx}
  \par\vspace{4pt}
  \begin{minipage}{\textwidth}
    \footnotesize
    \textit{Notes.} EM includes all attempts; S: scorable outputs; I/T: invalid/truncated counts. Leaf metrics are percentages macro-averaged over S. Tokens is the rounded median over completed responses. API failures are zero. Bold EM marks full correctness. Each row attempts 24 outputs; the two ID variants reuse source records and input permutations but were generated in separate batches.
  \end{minipage}
  \caption{\textbf{MBTA ordering validation with native and local trip IDs.} Only Base and Order belong to this comparison.}
  \label{tab:intervention-mbta-scores}
\end{table*}

\begin{table*}[t]
  \centering
  \footnotesize
  \setlength{\tabcolsep}{3pt}
  \renewcommand{\arraystretch}{1.10}
  \begin{tabularx}{\textwidth}{Xllrrrrrr}
    \toprule
    \textbf{Trip IDs} & \textbf{Model} & \textbf{Arm} & \textbf{N} & \textbf{Other} & \textbf{Outside} & \textbf{Absent} & \textbf{Conflict} & \textbf{Copies} \\
    \midrule
    \textbf{Native} & Qwen 9B & Base & 1104 & 9 & 4 & 2 & 8 & 77 \\
     &  & \textbf{Order} & 1152 & 0 & 0 & 0 & 0 & 0 \\
    \addlinespace[5pt]
     & Gemma 12B & Base & 1152 & 0 & 0 & 0 & 0 & 0 \\
     &  & \textbf{Order} & 1152 & 0 & 0 & 0 & 0 & 0 \\
    \addlinespace[5pt]
    \textbf{Local trip} & Qwen 9B & Base & 1152 & 3 & 23 & 23 & 10 & 261 \\
     &  & \textbf{Order} & 1152 & 0 & 0 & 0 & 0 & 0 \\
    \addlinespace[5pt]
     & Gemma 12B & Base & 1152 & 0 & 0 & 3 & 1 & 24 \\
     &  & \textbf{Order} & 1152 & 0 & 0 & 0 & 0 & 0 \\
    \bottomrule
  \end{tabularx}
  \par\vspace{4pt}
  \begin{minipage}{\textwidth}
    \footnotesize
    \textit{Notes.} N covers scorable reference records. Other/Outside/Absent/Conflict are mutually exclusive; Copies is separate. Source tuples are unique within each instance. Definitions and ambiguous-occurrence handling follow Appendix~\ref{app:intervention-protocol}.  Maximum N is 1,152 per row. The complete four-field tuples are unique within each source instance; this attribution uses a broader criterion than the strict unique-leaf movement analysis in Appendix~\ref{app:record-assignment}.
  \end{minipage}
  \caption{\textbf{MBTA record diagnostics.} Copies remain visible even when almost every target value is correctly bound.}
  \label{tab:intervention-mbta-records}
\end{table*}

\paragraph{Results, copies, and denominator.}
Base EM declines from \textbf{8/24 to 3/24 for Qwen} and from \textbf{24/24 to 17/24 for Gemma} with local trip IDs. Native-ID Qwen has one truncated output; the other cells are fully scorable. Order attains \textbf{24/24 for both models under both ID variants}, with no record errors, conflicts, or copies.

The additional failure forms include copied records that leave their correct target intact. Gemma's local-ID Base retains VPA 99.70\% but has precision 98.14\% and 24 extra complete-record copies. Across all uniquely attributable off-target prefix occurrences, 20 of its 24 occurrences retain the same trip label under a different route or direction; native-ID Gemma has none. For Qwen, total off-target prefix occurrences rise from 90 to 287, while these same-trip cross-parent occurrences rise from 9 to 125. They account for 116 of the net 197 additional occurrences. This aggregation includes movements and copies, rather than treating every extra declaration as a displaced source record.

All complete four-field source tuples are unique within each task, but individual values are often repeated. The tuple audit therefore keeps copies and conflicts visible without equating its broader attribution with the strict leaf tracing of Appendix~\ref{app:record-assignment}. An off-target prefix can preserve the expected container hierarchy while failing to match any valid reference record path; the Outside column alone does not identify a bracket-closure error. The MBTA evidence supports the observed ordering gains and path-confusion pattern within these controlled variants.

\subsection{Evidence and Reproduction Index}
\label{app:intervention-evidence}

The numbers above are reaggregated from saved per-output evaluations, with task hashes verified against the archived manifests and record diagnostics recomputed from retained declarations. No model is rerun and no raw response is rewritten. The experiment summaries and the following archive-relative run identifiers locate the evidence:

\begin{itemize}
  \item Fresh validation: \texttt{record-\allowbreak paths-\allowbreak replication-\allowbreak 20261007-\allowbreak 174436}, including the frozen \texttt{experiment-config.json} and separate model directories.
  \item Local shuffling: \texttt{prompt-\allowbreak layout-\allowbreak rerun-\allowbreak 20261005-\allowbreak 200348}, model subdirectories ending in \texttt{-pilot-order}.
  \item Large-model gradient: \texttt{pilot-scale-large-20261009}, including its frozen configuration and model directories.
  \item PDB: \texttt{pdb-\allowbreak stability-\allowbreak 20261008-\allowbreak 133133} for Qwen/Gemma, \texttt{pdb-\allowbreak ministral-\allowbreak 20261008-\allowbreak 140606} for Ministral, and \texttt{pdb-supplement-20261008} for the one-entry supplement.
  \item MBTA: \texttt{prompt-\allowbreak layout-\allowbreak rerun-\allowbreak 20261005-\allowbreak 200348} model directories ending in \texttt{-mbta}, and \texttt{gtfs-\allowbreak mbta-\allowbreak local-\allowbreak trip-\allowbreak 20261006-\allowbreak 120004} directories ending in \texttt{-mbta-local-trip}.
\end{itemize}

Runs reside under \texttt{JSONPlacement/data/runs}. PDB policies, exclusions, sequence audit, manifests, and combined results are under \texttt{JSONPlacement-data/real/pdb}; MBTA extracts and manifests are under \texttt{JSONPlacement-data/real/gtfs/mbta}. The generators in \texttt{src/dataset} are \texttt{replication.ts}, \texttt{order.ts}, \texttt{scale.ts}, \texttt{pdb.ts}, and \texttt{gtfs.ts}. Generation commits are \texttt{7006e62} (fresh validation), \texttt{530de92} (local shuffling), and \texttt{d50be7f} (large-model gradient). Dataset and per-task hashes in \texttt{run.json} identify the exact archived inputs.

\end{document}